\documentclass[10pt,twocolumn,letterpaper]{article}
\usepackage[dvipsnames, table]{xcolor}

\usepackage{iccv}              % To produce the CAMERA-READY version
\usepackage{amssymb}
\usepackage{booktabs}    
\usepackage{colortbl}  
\usepackage{graphicx}   
\usepackage{caption}    
\usepackage{pifont}
\usepackage{multirow}  
\usepackage{stfloats}
\usepackage{relsize}
\usepackage{mathtools}
\usepackage{threeparttable}
\usepackage{float}
\usepackage{placeins}
\usepackage{cuted} 
\usepackage{lineno}
\usepackage{multirow}            % for multirow in tables
\usepackage{makecell}            % for makecell
\usepackage[symbol]{footmisc}
\usepackage[accsupp]{axessibility}

\newcommand{\cmark}{\textcolor{ForestGreen}{\ding{51}}}%
\newcommand{\xmark}{\textcolor{BrickRed}{\ding{55}}}%

\definecolor{iccvblue}{rgb}{0.21,0.49,0.74}
\usepackage[pagebackref,breaklinks,colorlinks,allcolors=iccvblue]{hyperref}

\def\paperID{9789} % *** Enter the Paper ID here
\def\confName{ICCV}
\def\confYear{2025}

\title{\large{COIN: Confidence Score-Guided Distillation for Annotation-Free Cell Segmentation}\vspace{-0.2cm}}

\author{%
  Sanghyun Jo$^{1}$\footnotemark[1]
  \quad Seo Jin Lee$^{3}$\footnotemark[1]
  \quad Seungwoo Lee$^{1}$
  \quad Seohyung Hong$^{4}$
  \\[0.5ex]
  Hyungseok Seo$^{3}$\footnotemark[2]
  \quad Kyungsu Kim$^{2,4,5}$\footnotemark[2]
  \\[0.2ex]
  {\small\texttt{%
     \{shjo.april, vict.lee0\}@gmail.com \quad
     \{seojinleee, hong.sh, h.seo, kyskim\}@snu.ac.kr
  }}
}

\begin{document}
\maketitle
\begin{strip}
  \begin{center}
    \vspace{-2.2cm}
    {%
      {\small $^{1}$OGQ, Seoul, Korea}
      \vspace{-0.05cm}
      {\quad \small $^{2}$School of Transdisciplinary Innovations, Seoul National University, Korea}\\
      %\vspace{-0.1cm}
      {\small $^{3}$Laboratory of Cell \& Gene Therapy, Institute of Pharmaceutical Sciences, College of Pharmacy, Seoul National University, Korea}\\
      \vspace{-0.05cm}
      {\small $^{4}$Department of Biomedical Science and Medical Research Center, College of Medicine, Seoul National University, Korea} \\
      \vspace{-0.05cm}
      {\small $^{5}$Interdisciplinary Programs in Artificial Intelligence, Bioengineering, and Bioinformatics, Seoul National University, Korea}
    }
    \vspace{-0.2cm}
  \end{center}
\end{strip}

\footnotetext[1]{Equal contribution.}
\footnotetext[2]{Corresponding author.}

\begin{abstract}
Cell instance segmentation (CIS) is crucial for identifying individual cell morphologies in histopathological images, providing valuable insights for biological and medical research. While unsupervised CIS (UCIS) models aim to reduce the heavy reliance on labor-intensive image annotations, they fail to accurately capture cell boundaries, causing missed detections and poor performance. Recognizing the absence of error-free instances as a key limitation, we present COIN (\textbf{CO}nfidence score-guided \textbf{IN}stance distillation), a novel annotation-free framework with three key steps: (1) Increasing the sensitivity for the presence of error-free instances via unsupervised semantic segmentation with optimal transport, leveraging its ability to discriminate spatially minor instances, (2) Instance-level confidence scoring to measure the consistency between model prediction and refined mask and identify highly confident instances, offering an alternative to ground truth annotations, and (3) Progressive expansion of confidence with recursive self-distillation. Extensive experiments across six datasets show COIN outperforming existing UCIS methods, even surpassing semi- and weakly-supervised approaches across all metrics on the MoNuSeg and TNBC datasets. 

\small
\noindent\textbf{Project Page:} \href{https://shjo-april.github.io/COIN/}{\url{https://shjo-april.github.io/COIN/}}
% The code is available at: \href{https://github.com/shjo-april/COIN}{https://github.com/shjo-april/COIN}.
\vspace{-0.5cm}
\end{abstract}

\begin{figure}[t]
  \centering
  \includegraphics[width=1\linewidth]{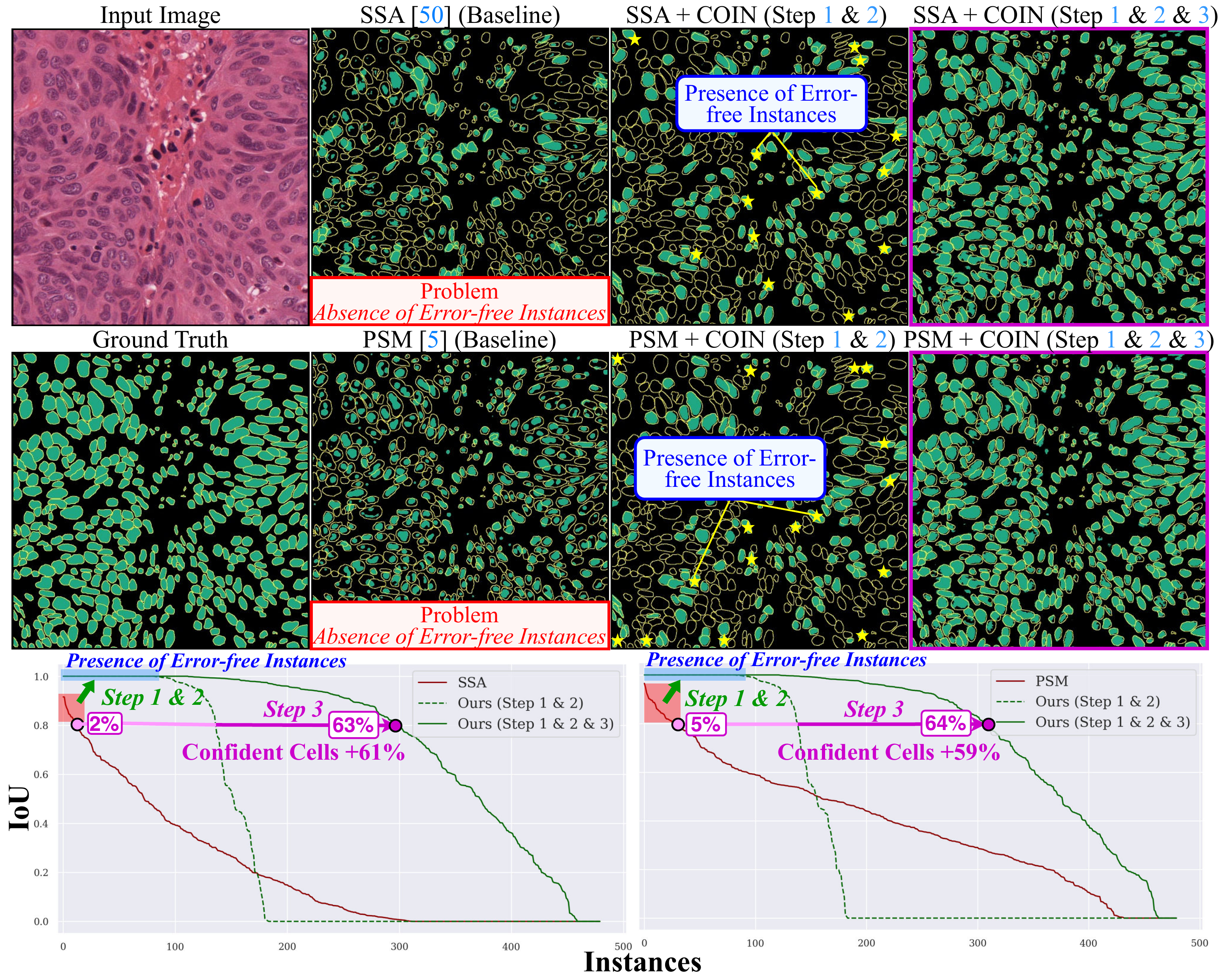}
  \vspace{-0.7cm}
  \caption{
      \textbf{Problem of \emph{the absence of error-free instances} with our solution.} Existing UCIS models \cite{ssa2020miccai, psm2023miccai} fail to produce error-free instances (\emph{i.e.}, IoU = 1). Our COIN recovers error-free instances (yellow stars) after \emph{Step 1} and \emph{Step 2}. Following this \emph{Step 3} increases the quantity and quality of confident cells (IoU $\geq$ 0.8).
    }
  \label{problem}
  \vspace{-0.6cm}
\end{figure}

\section{Introduction}
Mapping the precise locations and arrangements of individual cells in histopathological tissue images is essential to understand cellular interactions \cite{armingol2024nat} (\emph{e.g.,} ligand-receptor interactions) and to obtain diagnostic insights \cite{gurcan2009ieee, veta2014ieee, jothi2017review, jing2023compbio} (\emph{e.g.,} detection of cancer). However, cell instance segmentation (CIS) approaches \cite{graham2019hovernet, horst2024cellvit} involve time-consuming and labor-intensive pixel-level annotations \cite{jing2023compbio, he2022compbio}. Although weakly-supervised CIS (WCIS) methods \cite{xu2023sppnet, nam2024instasam} aim to reduce the annotation burden, they still rely on point/box annotations, {and unsupervised CIS (UCIS) methods fail to predict even a single instance with 100\% accuracy, a phenomenon we define as \emph{the absence of error-free instances} (\cref{problem}).}

%To completely eliminate labeling costs in CIS, recently developed unsupervised CIS (UCIS) methods \cite{carpenter2006cellprofiler, schindelin2012fiji, tian2020c2fnet} focus on extracting the most discriminative parts with geometric augmentation (\emph{e.g.}, rotation){, which fails to address edge-discriminative features. Therefore,} as shown in \cref{problem}, we observe that none of recent UCIS methods \cite{ssa2020miccai, psm2023miccai} could predict even a single instance with 100\% accuracy, a phenomenon we define as \emph{the absence of error-free instances}. 

To address this problem, we introduce COIN (\textbf{CO}nfidence score-guided \textbf{IN}stance distillation), a novel annotation-free, model-agnostic framework designed to automatically recover error-free instances and utilize them for confidence expansion. {Our research is grounded in two main hypotheses: \textbf{1) Limitation of Augmentation-based Learning:} Prior UCIS methods (\emph{e.g.,} PSM \cite{psm2023miccai}) rely heavily on geometric augmentations (\emph{e.g.,} rotation), which emphasize geometrically variant features (\emph{e.g.,} elongated shape) while neglecting subtle or round cells that appear unchanged. This bias results in incomplete instance masks (\cref{problem}). \textbf{2) Model-SAM Consistency as a Proxy for Error-free Instances:} Uniformly accepting all pseudo-labels propagates errors since it cannot differentiate reliable from noisy predictions. Thus, we propose an unsupervised accuracy measure based on the model's propagated masks and SAM-refined outputs, selecting only highly consistent instances (\emph{i.e.,} masks closely matching ground truth) for training (\cref{sam_ablation}).}

Our contributions are summarized as follows:

%refined masks from SAM \cite{kirillov2023segment}. \textbf{Third}, by iteratively scoring model predictions and distilling confident instances, our method iteratively distills these confident instances to expand the set of accurate instances (\cref{Recursive}). Our contributions are summarized as follows:
\begin{table*}[t]
  \caption{Conceptual comparison of semi- and weakly-supervised and unsupervised cell instance segmentation, and COIN (ours).} 
  \vspace{-0.3cm}
  \setlength{\tabcolsep}{3pt}
  \centering
  \begin{scriptsize}
  \begin{tabular}{p{0.4\textwidth} | c c c c | c c c}
      \toprule
      \textbf{Properties} & \multicolumn{4}{c|}{\cellcolor{gray!20}\textbf{Semi- and Weakly-supervised CIS}} 
                         & \multicolumn{3}{c}{\cellcolor{gray!20}\textbf{Unsupervised CIS}} \\  
      \cmidrule(lr){2-5} \cmidrule(lr){6-8}
      & \textbf{TextDiff} \cite{feng2024textdiff}
      & \textbf{SPPNet} \cite{xu2023sppnet} 
      & \textbf{InstaSAM} \cite{nam2024instasam} 
      & \textbf{UN-SAM} \cite{chen2024unsam} 
      & \textbf{SSA} \cite{ssa2020miccai} 
      & \textbf{PSM} \cite{psm2023miccai} 
      & \textbf{COIN (Ours)} \\  
      \midrule 
      Address cell instance segmentation & \cmark & \cmark & \cmark & \cmark & \cmark & \cmark & \cmark \\
      Do not require image-related annotations (\emph{e.g.}, point or box) for training & \xmark & \xmark & \xmark & \xmark & \cmark & \cmark & \cmark \\
      Propose instance (\emph{i.e.}, cell) propagation (\emph{e.g.}, USS+OT) & \xmark & \xmark & \xmark & \xmark & \xmark & \xmark & \cmark \\
      Propose instance-level (\emph{i.e.}, cell-level) scoring & \xmark & \xmark & \xmark & \xmark & \xmark & \xmark & \cmark \\
      Perform recursive learning & \xmark & \xmark & \cmark & \xmark & \xmark & \cmark & \cmark \\
      \bottomrule
    \end{tabular}
  \end{scriptsize}
  \vspace{-0.5cm}
  \label{contributions}
\end{table*}

\begin{itemize}
    \item We recognize and overcome a challenge inherent in existing UCIS models \cite{ssa2020miccai,psm2023miccai}, which we refer to as \emph{the absence of error-free instances}.
    \item We propose COIN (\textbf{CO}nfidence score-guided \textbf{IN}stance distillation), featuring a novel instance scoring approach that alleviates the dependence on ground truth annotations, followed by expansion of confidence via recursive self-distillation that progressively increases the number of highly confident instances.
    \item {COIN establishes new state-of-the-art performance on the MoNuSeg and TNBC datasets with at least a 9\%p improvement over previous UCIS methods \cite{ssa2020miccai,psm2023miccai} on the MoNuSeg test set, and it even surpasses semi- and weakly-supervised approaches \cite{liu2022weakly,nam2024instasam,feng2024textdiff} (\cref{main_table}).}
\end{itemize}

\section{Related Work}
\subsection{Unsupervised Instance Segmentation}
Advances in unsupervised instance segmentation (UIS) \cite{li2024promerge, wang2023cutler} have provided useful insights into computer vision, yet their direct application to histopathology remains suboptimal. Unlike natural images with a clear distinction between objects and background, histological images contain densely packed, morphologically diverse cells without clear boundaries. As a result, UIS models tend to merge adjacent nuclei or fragment individual cells due to texture variations \cite{xie2024unsupervised}. Also, all existing UIS models depend on the USS backbone and fix the number of instances within an image, resulting in lower performance on cell benchmarks that require detecting hundreds of instances (see \cref{main_table}), highlighting the need for cell instance segmentation (CIS) methods specifically designed for histopathological images.
%Methods like MaskCut and CutLER \cite{wang2023cutler} leverage self-supervised features (\emph{e.g.,} DINO ViT) to segment objects in natural images, while ProMerge \cite{li2024promerge} refines mask groupings.

\subsection{Annotation-Driven Cell Segmentation}
Recently, many semi- and weakly-supervised cell segmentation models \cite{feng2024textdiff, xu2023sppnet, nam2024instasam, hsu2019bbtp, wang2021bbwsis, tian2020boxinst, cui2023allinsam, wang2024mudslide, lee2023mediar, wu2022cross} have been developed to alleviate the need for manual annotations. For supervision, recent methods \cite{xu2023sppnet, nam2024instasam} utilize Segment Anything Model (SAM) \cite{kirillov2023segment} as their foundational baseline, which requires manual human prompting (\emph{e.g.}, points) \cite{xu2023sppnet, nam2024instasam, cui2023allinsam}.

{Although previous SAM-based methods \cite{xu2023sppnet, nam2024instasam} attempt to diminish the burden of manual pixel-level annotations by using weak labels (\emph{e.g.}, box per cell), they still rely on SAM outputs generated from manually annotated boxes or points. In contrast, we leverage SAM without relying on any annotations and achieve unsupervised instance scoring for the first time. In this paper, we propose a novel instance scoring to measure the consistency between model predictions and pseudo-GT masks using SAM (\cref{step2}), alleviating the reliance on manual instance-level annotations (\emph{e.g.,} cells).

\subsection{Annotation-Free Cell Segmentation}
To tackle the shortcomings of weakly-supervised cell segmentation, some models focus on unsupervised approaches that do not require any image-related annotations \cite{schmidt2018stardist, han2021usar, bescond2022sop, zhao2023cncsa, wolf2023cellulus, yang2024hvsunsup, farhad2018dcgmm, nan2022dcgn, nan2024damm, carpenter2006cellprofiler, schindelin2012fiji, hou2019ieee}. USAR \cite{han2021usar} is based on the adversarial learning paradigm that incorporates U-net structure, and the method proposed by Bescond \emph{et al.} \cite{bescond2022sop} leverages a discriminator that is trained on segmentation mask features from the public dataset. As for UCIS approaches, SSA \cite{ssa2020miccai} adopts scale classification as a self-supervision signal, assuming that the nuclei size and texture can indicate the magnification level. Prior self-activation map (PSM) \cite{psm2023miccai} employs rotation as the mechanism to extract variant features of the morphology of the cell and invariant features of tissue for self-supervised learning. 

These approaches, however, are highly error-prone since their prediction relies on partial cell features and thus fails to achieve a complete instance of the cell (\cref{problem}). To the best of our knowledge, our method is the first to address \emph{the absence of error-free instances} and thereby substantially enhance cell segmentation sensitivity without relying on image-related annotation (\cref{contributions}). 

\begin{figure*}[htbp]
  \centering
  \includegraphics[width=0.9\textwidth]{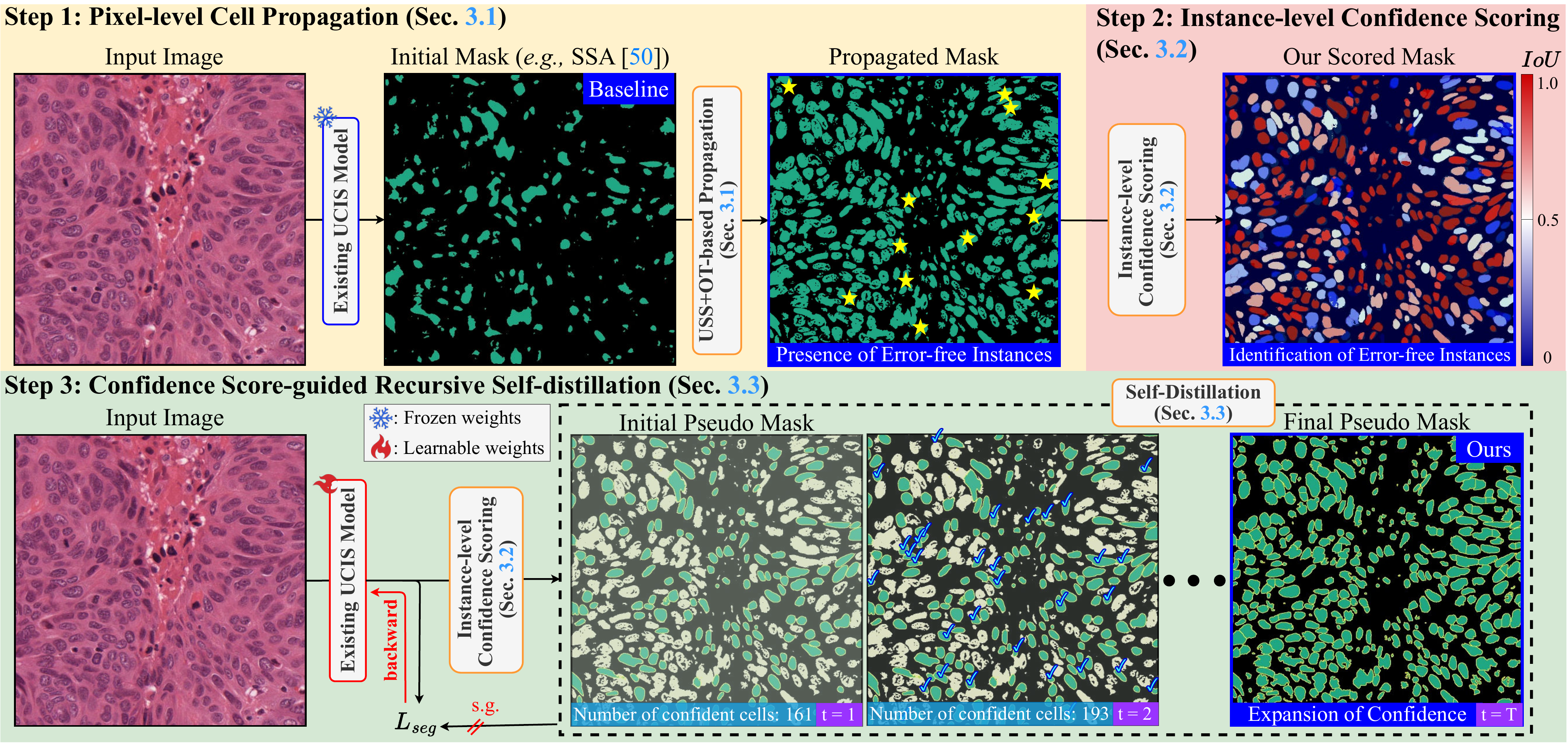}
  \vspace{-0.2cm}
  \caption{ \textbf{Overview of COIN.} Our approach is divided into three stages. Images with blue boundaries indicate the output of each step.}
  \label{overview}
  \vspace{-0.5cm}
\end{figure*}

\section{Method}
COIN is divided into three steps: (1) Pixel-level cell propagation (\cref{step1}), (2) Instance-level confidence scoring (\cref{step2}), and (3) Confidence score-guided recursive self-distillation (\cref{step3}). Our framework is depicted in \cref{overview}.

\begin{figure}[t]
  \centering
  \includegraphics[width=1.04\linewidth]{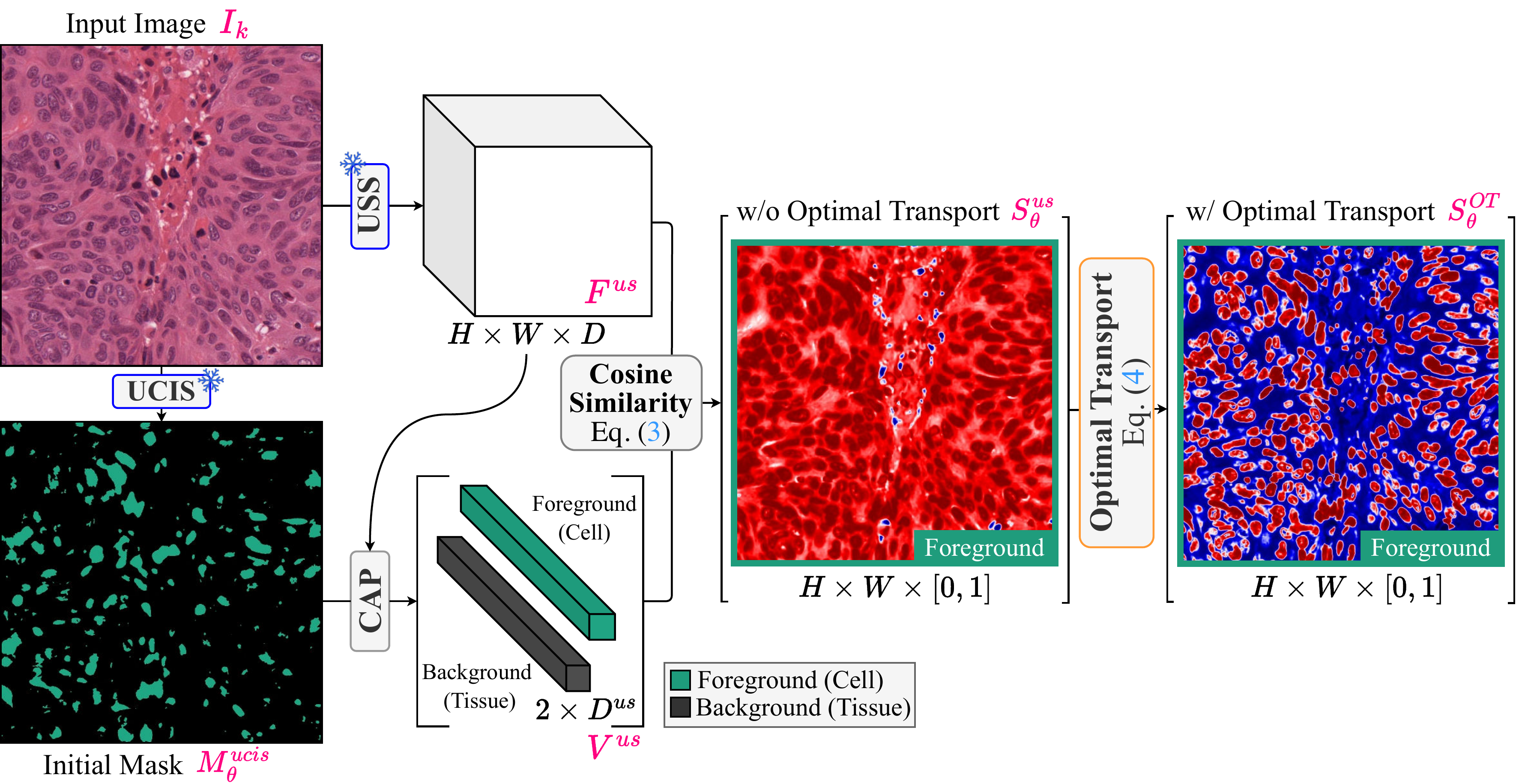}
  \vspace{-0.7cm}
  \caption{\textbf{Illustration of pixel-level cell propagation.} The USS centroid and the embedding vectors from SSA \cite{ssa2020miccai} are clustered, generating a similarity map. The application of OT refines the overlapping activation, allowing for a clearer distinction between foreground and background.}
  \vspace{-0.6cm}
  \label{USS+OT}
\end{figure}

\subsection{\label{step1}(Step 1) Pixel-level cell propagation}
In \emph{Step 1}, to guarantee the presence of an error-free instance, our proposed method begins with cell propagation by unsupervised semantic segmentation (USS) \cite{caron2021emerging, oquab2023dinov2, he2022masked} on the $k^{th}$ input histopathological image $I_k \in \mathbb{R}^{ H \times W \times C}$, in which $H$ refers to the height, $W$ refers to the width, and $C$ indicates the number of channels in the input image (\emph{i.e.,} 3 for RGB images). Then, the refinement by optimal transport (OT) \cite{rachev1985monge} follows, as summarized below:
\begin{align} \label{eq_step1} 
M^{ucis}_{\theta}(I_k) \xrightarrow[\text{USS}]{\ } S^{us}_{\theta}(I_k) \xrightarrow[\text{OT}]{\ } S^{OT}_{\theta}(I_k)
\end{align}
Initially, we obtain unsupervised feature map denoted as
\begin{align} \label{eq_F}
F^{us} = U(I_k) \in \mathbb{R}^{ H \times W \times D^{us}}
\end{align}
from $I_k$. Here, $U$ refers to the unsupervised encoder (\emph{e.g.,} DINOv2 \cite{oquab2023dinov2} and MAE \cite{he2022masked}), and the superscript $us$ indicates the unsupervised component.

At the same time, the initial mask $M^{ucis}_{{\theta}}(I_k) \in \mathbb{R}^{ H \times W}$ is acquired from existing UCIS model (\emph{e.g.,} SSA \cite{ssa2020miccai}), which takes $I_k$ and predicts cell regions (\cref{USS+OT}). The initial value for $\theta$ is the parameter of existing pre-trained UCIS models (\emph{i.e.,} $\theta_{init}$), which gets updated by our proposed method, thereby consistently improving the baseline UCIS model. For more details on USS, refer to \cref{appendix_uss_details}.
 
We apply class-level average pooling (CAP) \cite{jo2024dhr} to the initial mask $M^{ucis}_{{\theta}}(I_k)$, in which the embedding vectors for each class (cell or tissue) are grouped based on the mask and computes the average of the vectors assigned to the same group (see \cref{appendix_cap} for more details on CAP). This creates USS centroids for each class, denoted as $V^{us}_{\theta} = CAP(F^{us}, M^{ucis}_{{\theta}}(I_k)) \in \mathbb{R}^{ 2 \times D^{us}}$. 

\begin{figure}[t]
  \centering
  \includegraphics[width=0.9\linewidth]{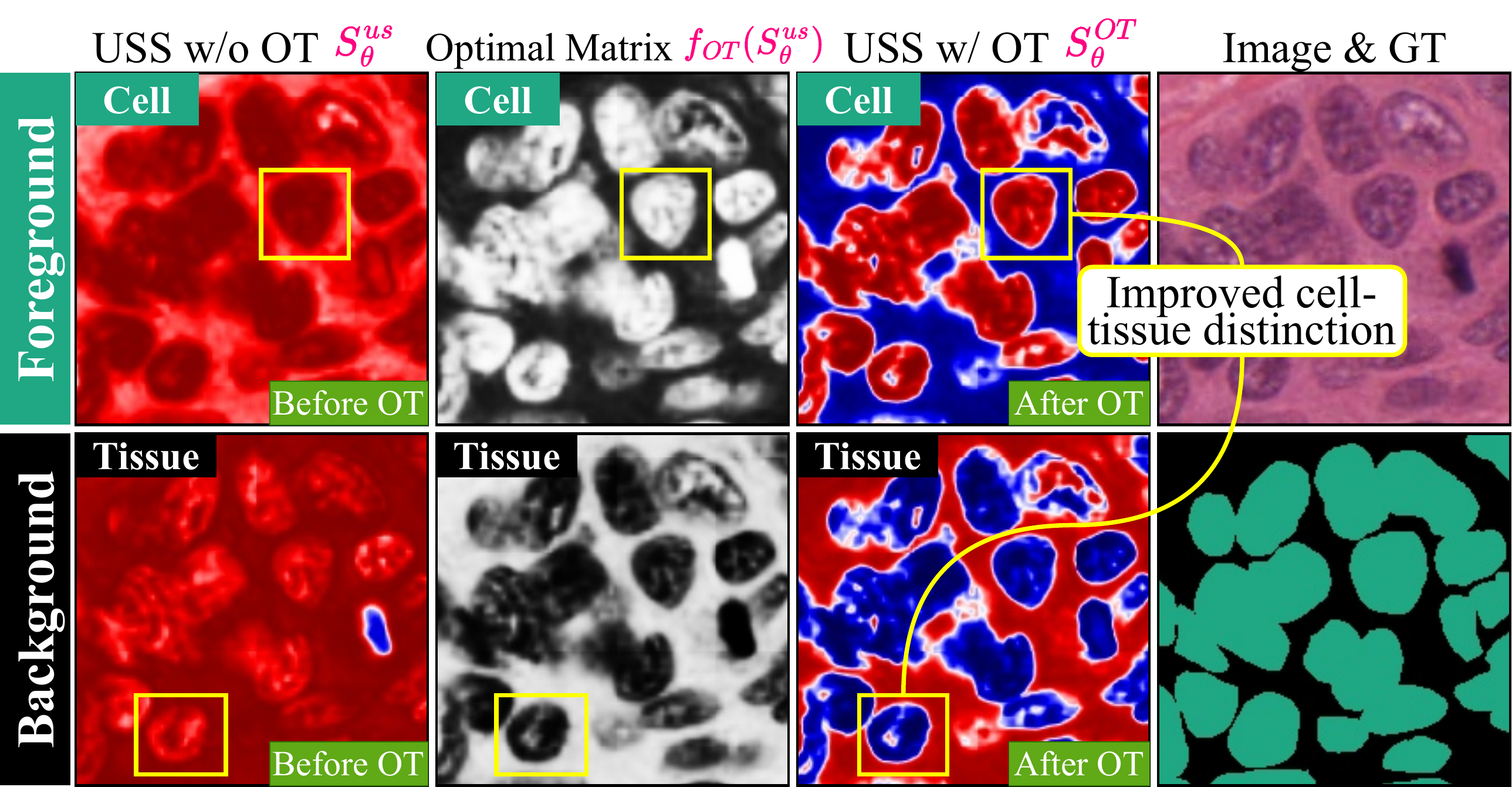}
  \vspace{-0.2cm}
  \caption{\textbf{Visualization of the effect of OT.} In \cref{eq_S_OT}, multiplying the similarity map $S_{\theta}^{us}$ with the optimal matrix $f_{OT}(S_{\theta}^{us})$ yields a non-overlapping mask $S_{\theta}^{OT}$ that better preserves instance boundaries.}
  \label{OT_effect}
  \vspace{-0.5cm}
\end{figure}

From here, $F^{us}$ and $V^{us}_{\theta}$ are clustered, producing similarity map, which is written as 
\begin{align} \label{eq_S_us}
S^{us}_{\theta}(I_k) := ReLU(sim(F^{us}_{ij}, V^{us}_{\theta}))\in \mathbb{R}^{H \times W \times [0,1]},
\end{align}
with $sim(\cdot)$ denoting pixel-level cosine similarity and $ij$ referring to 2D coordinates. 

The USS model (\emph{e.g.,} MAE \cite{he2022masked}) employed in our method is trained with natural images and not histopathological tissue images, causing an elevated rate of false positives. These models tend to distinguish objects based on color \cite{jo2024dhr}, and their pixel similarity-based grouping often fails to distinguish between cells of similar colors. As shown in \cref{OT_effect}, the USS output fails to differentiate the activation pattern of the cell from the background. We address this substantial increase in FP by incorporating OT, which identifies an optimal matrix where each pixel (\emph{e.g.}, cell) in the similarity map is assigned optimally to prevent overlap between the pixels. This can result in high matrix values even when there is minimal cell prediction, leading to potential false positives \cite{jo2024dhr}. To address this, we multiplied the results in \cref{eq_S_OT} to filter out consistent pixels. Importantly, OT’s emphasis on the minor class (\emph{e.g.,} cells with fewer pixels) reduces FN in cell regions \cite{jo2024dhr}. As depicted in \cref{OT_effect}, the USS-predicted cell boundaries are not well-defined, but they become more distinguished after OT is applied. We selected OT for its outstanding performance compared to other clustering methods \cite{neal1998gmm, macqueen1967kmeans}, as other methods fail to distinguish overlapping pixels (see \cref{ot_alternatives} for details). % We selected OT for its outstanding performance compared to other clustering methods \cite{neal1998gmm, macqueen1967kmeans} (see \cref{ot_alternatives}).}

As indicated in \cref{eq_step1}, the application of OT further refines pseudo labels as
{\small
\begin{align} \label{eq_S_OT}
S^{OT}_{{\theta}}(I_k) = f_{OT}(S^{us}_{\theta}(I_k))\cdot S^{us}_{\theta}(I_k) \in \mathbb{R}^{H \times W \times [0,1]},
\end{align}
}
where {\smaller {${f_{OT}(S^{us}_{\theta}(I_k))} := \underset{T}{\arg\min}\sum_{i=1}^{HW} \sum_{j=1}^{C} T_{ij} \left(1 - S^{us}_{ij}\right) - \lambda H(T)$}}. Here, $S^{OT}_{{\theta}}(I_k)$ has dimension $H \times W \times [0, 1]$ with the number 1 denoting foreground (\emph{i.e.,} cell) and the number 0 denoting background (\emph{e.g.,} tissue). $T$ indicates the OT matrix, which performs OT-based propagation on the input similarity map $S^{us}_{\theta}(I_k)$ {(see \cref{rebutt_push_operation} for more details)}. $\lambda$ is the regularization parameter set to 0.1, and H($\cdot$) refers to the entropy term \cite{jo2024dhr}. The refined outcome $S^{OT}_{{\theta}}(I_k)$ is the final propagated mask of the first step. To summarize, \emph{Step 1} allows the detection of all cells with high sensitivity and ensures the presence of an error-free instance. 

\begin{figure*}[t]
  \centering
  \includegraphics[width=0.9\linewidth]{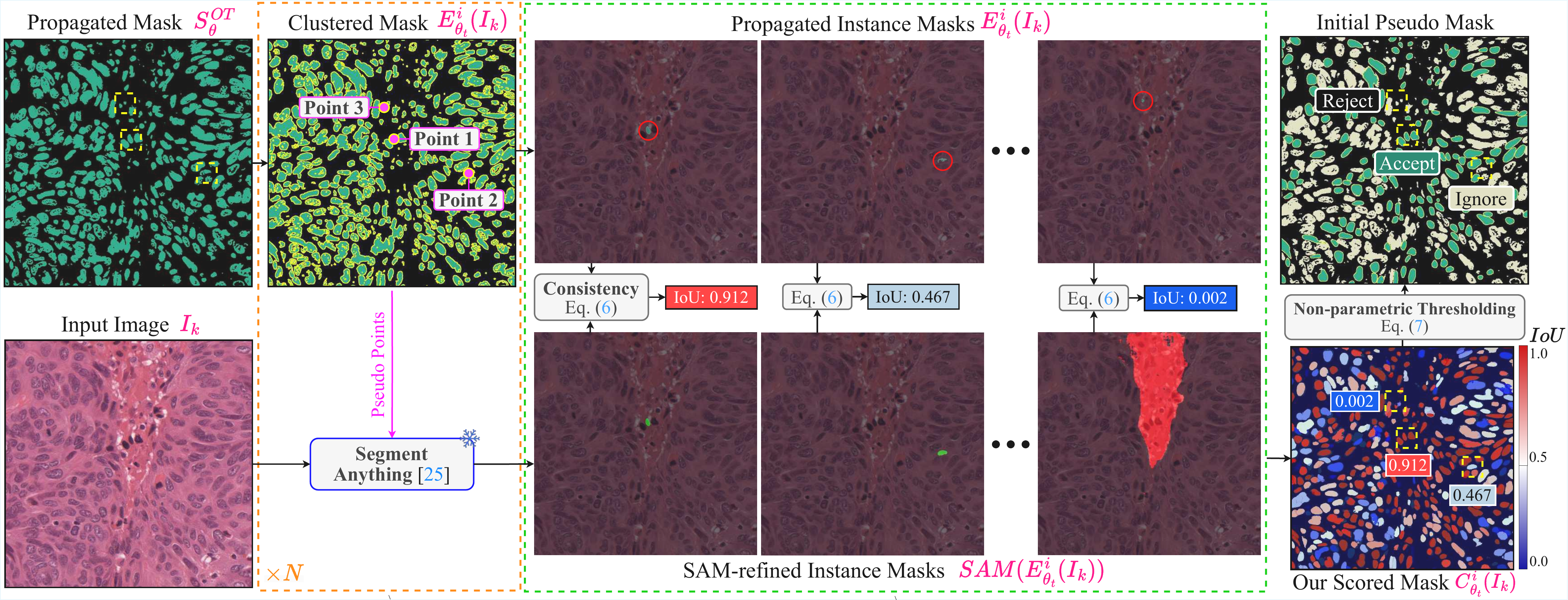}
  \vspace{-0.3cm}
  \caption{\textbf{Illustration of instance-level confidence scoring.} The center point of the instances obtained from the propagated mask is used as the point prompt for SAM \cite{kirillov2023segment}. The consistency between our propagated mask and SAM-generated pseudo-GT mask is measured, and only those above the non-parametric threshold $\delta_k$ are accepted.}
  \label{InstanceScoring}
  \vspace{-0.35cm}
\end{figure*}

\subsection{\label{step2}(Step 2) Instance-level Confidence Scoring}
In \emph{Step 2}, we assess the confidence of each instance by applying an unsupervised scoring approach to extract error-free instances. Without relying on the actual ground truth (GT) annotations (\emph{i.e.,} manual pixel-level masks), this step scores each instance based on its consistency with the pseudo-GT mask, thereby spotting error-free instances that have AJI scores close to 1 (see \cref{figure_correlation}). The general mechanism of \emph{Step 2} is summarized below:
{\small
\begin{align} \label{eq_step2}
S^{OT}_{\theta}(I_k) \xrightarrow[\text{Watershed}]{\,} \left\{ E^i_{\theta_t}(I_k) \right\}_{i=1}^{N} \xrightarrow[\text{Scoring}]{\,} \left\{ C^i_{\theta_t}(I_k) \right\}_{i=1}^{N}
\end{align}
}
Using instance clustering approaches (\emph{i.e.,} CCL \cite{rosenfeld1966sequential}), we generate $N$ instance masks. Then, adopting the methodology in previous UCIS study \cite{ssa2020miccai}, we separate the instances denoted as $E^i_{{\theta_t}}(I_k)$ with the watershed algorithm (\cref{appendix_watershed}). From here, the center point of $E^i_{{\theta_t}}(I_k)$ is utilized as the point prompt for SAM \cite{kirillov2023segment}, generating a pseudo-GT mask $SAM(E^i_{{\theta}_t}(I_k))$ (\cref{InstanceScoring} and \cref{appendix_scoring}).  
%To separate total $N$ instances denoted as $E^i_{{\theta_t}}(I_k)$, the propagated mask $S^{OT}_{{\theta}}(I_k)$ obtained from the previous step is applied to watershed algorithm \cite{ssa2020miccai}(see \cref{rebutt_comp_ablation} and \cref{appendix_watershed} for more details). Then, the center point of $E^i_{{\theta_t}}(I_k)$ is utilized as the point prompt for SAM \cite{kirillov2023sam}, generating pseudo-GT mask $SAM(E^i_{{\theta}_t}(I_k))$ (\cref{InstanceScoring} and \cref{appendix_scoring}).  

% \begin{figure}[t]
%   \centering
%   \includegraphics[width=1.02\linewidth]{ICCV 2025/figures_main/Figure4_Scoring.drawio.pdf}
%   \vspace{-0.8cm}
%   \caption{\textbf{Illustration of instance-level confidence scoring.} The center point of the instances obtained from the propagated mask is used as the point prompt for SAM \cite{kirillov2023sam}. The consistency between our propagated mask and SAM-generated pseudo-GT mask is measured, and only those above the non-parametric threshold $\delta_k$ are accepted.}
%   \label{InstanceScoring}
%   \vspace{-0.3cm}
% \end{figure}

Instead of directly using SAM-generated pseudo-GT masks, we take a step to compare them to the model prediction because SAM working alone often results in faulty predictions due to prompt sensitivity (see \cref{InstanceScoring} and \cref{appendix_SAM_fig}). In other words, if the point prompt targets the background (\emph{e.g.,} tissue), SAM struggles to accurately differentiate between pixels, resulting in substantial classification errors. We solve this problem by cross-checking the model prediction $E^i_{{\theta_t}}(I_k)$ with the SAM-derived mask $SAM(E^i_{{\theta_t}}(I_k))$, and retaining only the instances that are consistent in both masks for training. In the later section, we confirm that our proposed consistency comparison allows the selection of confident instances close to GT, meaning AJI scores close to 1, which random sampling of instances fails to achieve (see \cref{figure_correlation} and \cref{appendix_sam}). 

Our proposed confidence scoring occurs for each instance \( i = 1, 2, \dots, N \) based on the consistency between our propagated mask and pseudo-GT mask, which can be defined as 
\begin{align} \label{eq_C}
C^i_{{\theta_t}}(I_k) = IoU(E^i_{{\theta_t}}(I_k), SAM(E^i_{{\theta_t}}(I_k))). 
\end{align}
After consistency comparison, the initial propagated masks are generated as follows:
\begin{align} \label{eq_decision}
\hat{M}_{k}^i =
\begin{cases}
1, & \text{if } C^i_{{\theta_t}}(I_k) > \delta_k, \\
-1, & \text{if } C^i_{{\theta_t}}(I_k) \leq \delta_k \text{ and } C^i_{{\theta_t}}(I_k) > 0, \\
0, & \text{if } C^i_{{\theta_t}}(I_k) = 0.
\end{cases}
\end{align}
Here, the non-parametric threshold $\delta_k$ is the sum of the mean and standard deviation of the confidence scores of all instances in the $k^{th}$ sample. COIN considers both the background ($\hat M_{k}^i = 0$)  and cells ($\hat M_{k}^i = 1$)  as certain, while the rest ($\hat M_{k}^i = -1$) are rejected and omitted from training  (\cref{InstanceScoring}). We define the set of accepted indices as 
\begin{align} \label{eq_A}
\mathcal{A}_{\delta} :=  \{i \, |\hat{M}_{k}^i = 1 {\text{ or }} \hat{M}_{k}^i = 0 \}. 
\end{align}

\begin{figure}[t]
  \centering
  \includegraphics[width=0.9\linewidth]{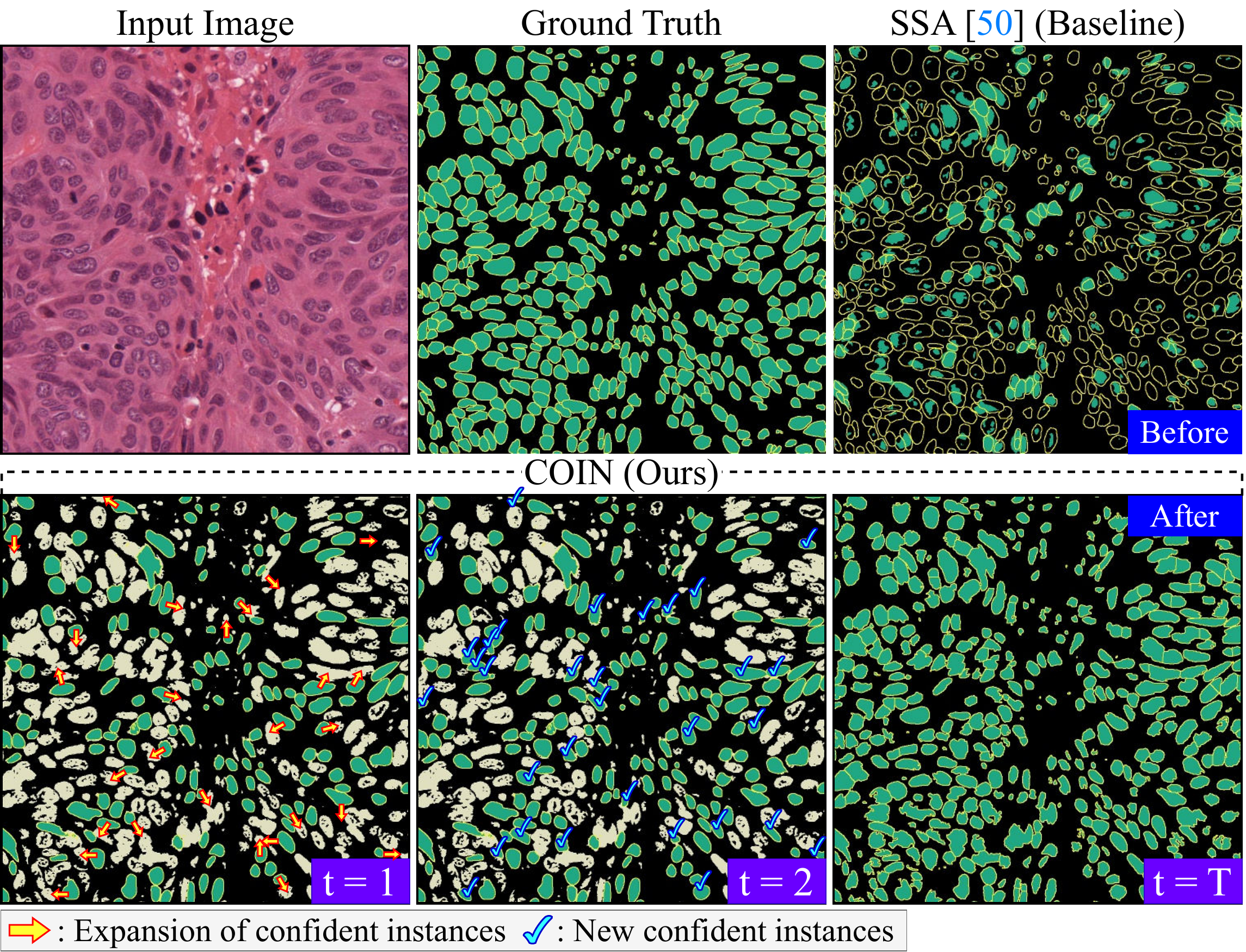}
  \vspace{-0.2cm}
  \caption{\textbf{Illustration of confidence score-guided recursive self-distillation.} The pseudo mask located at the bottom left (t=1) is the combined representation of the binary and edge pseudo masks. Confident instances with high scores are used for self-distillation, as indicated by yellow arrows. Each iteration results in more confident instances (\emph{e.g.,} blue check marks at t=2).}
  \label{Recursive}
  \vspace{-0.5cm}
\end{figure}

\subsection{\label{step3}(Step 3) Confidence Score-guided Recursive Self-distillation}

In \emph{Step 3}, recursive self-distillation occurs with the guidance from $\mathcal{A}_{\delta}$, allowing progressive increment in the number of confident cells. As the parameter $\theta_t$ gets altered from training, the accepted indices in \cref{eq_A} also change, allowing dynamic regulation of highly confident instances. For this, our method utilizes two pseudo masks for training (see \cref{appendix_edge}). 

To begin with, the pseudo binary and edge masks of accepted instances are denoted as
\begin{align} \label{eq_bin}
\hat{M}_{bin}^i(t) =  \left\{ E^i_{{\theta_t}}(I) \right\}_{i \in \mathcal{A}_{\delta}} \in  \{0, 1\}^{H \times W}, 
\\
\hat{M}_{edge}^i(t) =  \left\{ Edge(E^i_{{\theta_t}}(I)) \right\}_{i \in \mathcal{A}_{\delta}}\in  \{0, 1\}^{H \times W}, \label{eq_edge}
\end{align}
respectively. For the edge mask, $Edge(\cdot)$ indicates the traditional edge algorithm (\emph{i.e.,} Canny \cite{canny1986computational}). Our method includes an edge decoder for cell boundaries, which is not present in existing UCIS models, to discriminate instances of adjacent cells (see \cref{appendix_edge}, \ref{appendix_edgedecoder}, and \ref{appendix_adjacent} for more details on the edge decoder).

Lastly, COIN undergoes recursive self-distillation by utilizing pseudo masks $\hat{M}_{bin}^i(t)$ and $\hat{M}_{edge}^i(t)$. The total loss used in recursive self-distillation can be denoted as
{\scriptsize
\begin{align} \label{eq_loss}
\mathcal{L}(t) = 
 \mathcal{L}_{seg}(M^{ucis}_{bin}(I_{k;\theta_t}), \hat{M}_{bin}^i(t)) + \mathcal{L}_{seg}(M^{ucis}_{edge}(I_{k;\theta_t}), \hat{M}_{edge}^i(t)).
\end{align}
}
Here, $\mathcal{L}_{seg} = \mathcal{L}_{ce} + \mathcal{L}_{dice}$, in which $\mathcal{L}_{ce}$ is the cross-entropy loss, and $\mathcal{L}_{dice}$ refers to the Dice loss. Inspired by previous studies \cite{graham2019hovernet, xier2020seannet}, we distinguish $ \mathcal{L}_{seg}$ for each pseudo mask from \cref{eq_bin} and \cref{eq_edge} to specify instance-level learning.

The recursive self-distillation training with these losses progressively enhances the quality of initial masks generated by SSA \cite{ssa2020miccai} and increases the number of confident cells, as depicted in \cref{Recursive}.

% 241112 updated
\begin{table*}[htbp]
    \centering
    \caption{Performance comparison on the MoNuSeg \cite{kumar2017monuseg, kumar2020monuseg} and TNBC \cite{naylor2019tnbc} test sets.}
    \vspace{-0.3cm}
    \label{main_table}
    \scriptsize
    \setlength{\tabcolsep}{6pt}
    \renewcommand{\arraystretch}{0.85}
    \begin{threeparttable}
    \begin{tabular}{p{5cm}c|cccc|cccc}
        \toprule
        \textbf{Method} & \textbf{Cell Supervision} & \multicolumn{4}{c|}{\textbf{MoNuSeg}} & \multicolumn{4}{c}{\textbf{TNBC}} \\
        \cmidrule(lr){3-6} \cmidrule(lr){7-10}
         &  & \textbf{AJI} ($\uparrow$) & \textbf{PQ} ($\uparrow$) & \textbf{IoU} ($\uparrow$) & \textbf{Dice} ($\uparrow$) & \textbf{AJI} ($\uparrow$) & \textbf{PQ} ($\uparrow$) & \textbf{IoU} ($\uparrow$) & \textbf{Dice} ($\uparrow$) \\
        \midrule
        \multicolumn{10}{l}{\textbf{Annotation-free Instance Segmentation}} \\
        MaskCut \cite{wang2023cutler} {\tiny CVPR'23} & \xmark & 0.001* & 0.000* & 0.053* & 0.089* & 0.006* & 0.000* & 0.048* & 0.088* \\
        CutLER \cite{wang2023cutler} {\tiny CVPR'23} & \xmark & 0.002* & 0.000* & 0.143* & 0.244* & 0.003* & 0.000* & 0.082* & 0.146* \\
        ProMerge \cite{li2024promerge} {\tiny ECCV'24} & \xmark & 0.000* & 0.000* & 0.013* & 0.024* & 0.004* & 0.000* & 0.046* & 0.076* \\
        \midrule
        \multicolumn{10}{l}{\textbf{Annotation-free Cell Instance Segmentation}} \\
        CellProfiler \cite{carpenter2006cellprofiler} {\tiny Genome Biology'06} & \xmark & 0.123 & - & - & 0.404 & 0.208 & - & - & 0.415 \\
        Fiji \cite{schindelin2012fiji} {\tiny Nature Methods'12} & \xmark & 0.273 & - & - & 0.665 & - & - & - & - \\
        Hou \emph{et al.} \cite{hou2019ieee} {\tiny CVPR'19} & \xmark & 0.498 & - & - & 0.750 & - & - & - & - \\
        SSA \cite{ssa2020miccai} {\tiny MICCAI'20} & \xmark & 0.259* & 0.185* & 0.618* & 0.575* & 0.273* & 0.253* & 0.647* & 0.538* \\
        \rowcolor{gray!20}SSA + COIN (Ours) & \xmark & \textbf{0.580} & \textbf{0.536} & \textbf{0.776} & \textbf{0.794} & \textbf{0.568} & \textbf{0.540} & \textbf{0.797} & \textbf{0.774} \\
        PSM \cite{psm2023miccai} {\tiny MICCAI'23} & \xmark & 0.471* & 0.355* & 0.689* & 0.682* & - & - & - & - \\
        \rowcolor{gray!20}PSM + COIN (Ours) & \xmark & \textbf{0.579} & \textbf{0.539} & \textbf{0.777} & \textbf{0.797} & - & - & - & - \\
        \midrule
        \multicolumn{10}{l}{\textbf{Weakly-supervised Cell Instance Segmentation}} \\
        Qu \emph{et al.} \cite{qu2019weakly} {\tiny MIDL'19} & Point & 0.496 & - & - & 0.702 & - & - & - & - \\
        C2FNet \cite{tian2020c2fnet} {\tiny MICCAI'20} & Point & 0.493 & - & 0.624 & - & - & - & - & - \\
        Mixed Anno \cite{qu2020nuclei} {\tiny ISBI'20} & Point \& Mask & 0.516 & - & - & 0.733 & - & - & - & - \\
        BB-WSIS \cite{wang2021bbwsis} {\tiny MICCAI'21} & Box & - & - & - & 0.728 & - & - & - & 0.703 \\
        Liu \emph{et al.} \cite{liu2022weakly} {\tiny ISBI'22} & Point & 0.534 & - & - & 0.740 & - & - & - & - \\
        SPPNet \cite{xu2023sppnet} {\tiny MLMI'23} & Point & 0.497* & 0.392* & 0.709* & 0.719* & - & - & - & - \\
        All-in-SAM \cite{cui2023allinsam} {\tiny IOPscience'23} & Box & 0.502 & - & - & 0.738 & - & - & - & - \\
        PROnet \cite{nam2023pronet} {\tiny MICCAI'23} & Point & 0.555 & - & - & 0.750 & - & - & - & - \\
        InstaSAM \cite{nam2024instasam} {\tiny MICCAI'24} & Point & 0.574 & - & - & 0.772 & - & - & - & - \\
        \midrule
        \multicolumn{10}{l}{\textbf{Semi-supervised Cell Instance Segmentation}} \\
        CDCL \cite{wu2022cross} {\tiny CVPR'22} & Mask & - & - & - & 0.782 & - & - & - & - \\
        TextDiff \cite{feng2024textdiff} {\tiny MICCAI'24} & Mask \& Text & 0.510* & 0.410* & 0.726* & 0.726* & 0.464* & 0.358* & 0.728* & 0.666* \\
        \bottomrule
    \end{tabular}
    \begin{tablenotes}
        \item *: Directly reproduced results using publicly accessible code for a fair comparison. The rest are the values reported in the publication.
        \item -: Values not publicly disclosed.
    \end{tablenotes}
    \end{threeparttable}
    \vspace{-0.5cm}
\end{table*}

% 241108 updated
\begin{figure*}[t]
  \centering
  \includegraphics[width=1.0\textwidth]{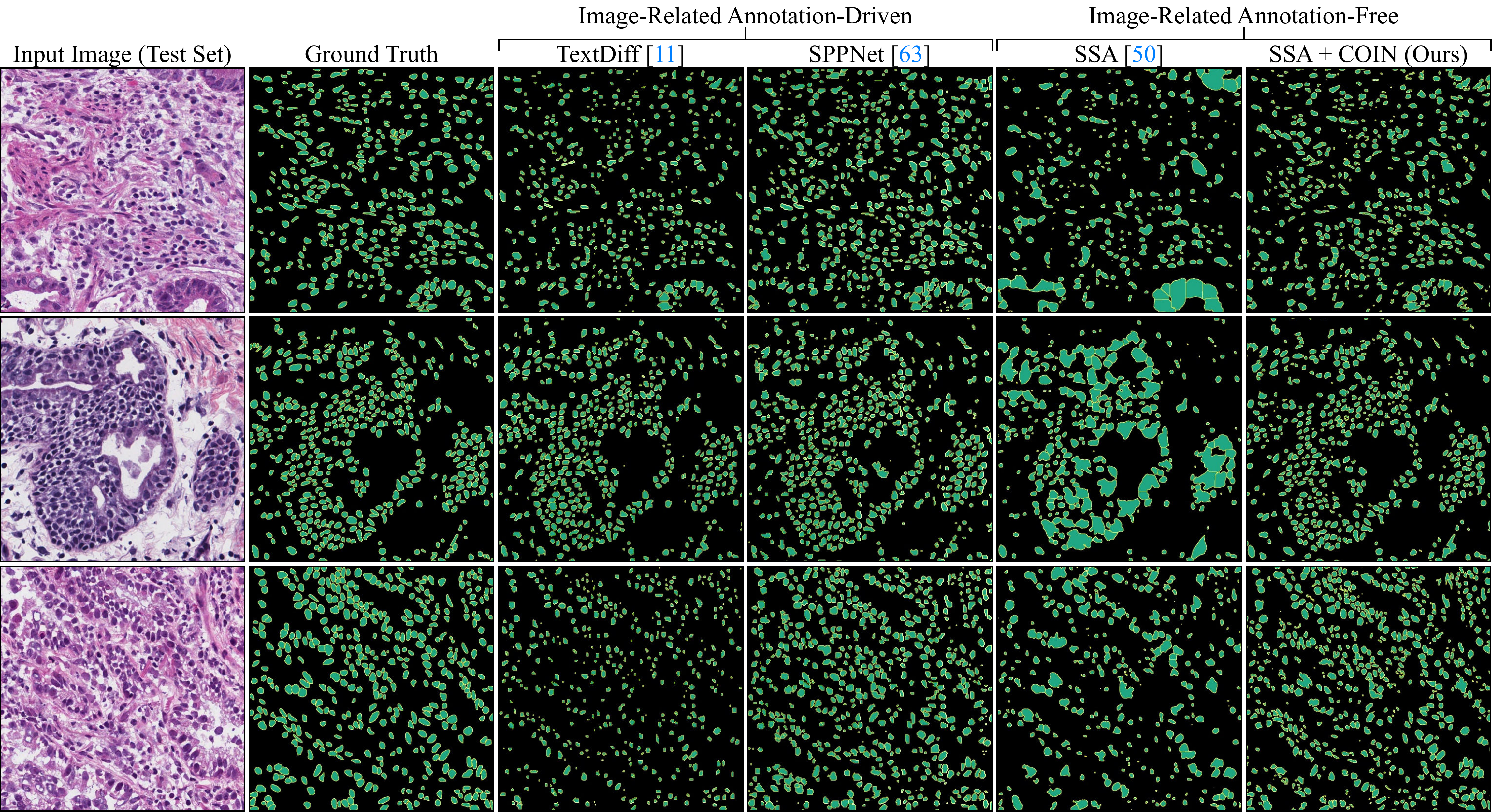}
  \vspace{-0.7cm}
  \caption{Qualitative comparison of image-related annotation-driven and -free methods \cite{ssa2020miccai, feng2024textdiff, xu2023sppnet} on MoNuSeg \cite{kumar2017monuseg, kumar2020monuseg} test set.}
  \label{main_results}
  \vspace{-0.4cm}
\end{figure*}

\vspace{-0.1cm}
\section{Experiments}

\subsection{\label{setup}Experimental Setup}
\vspace{-0.1cm}
\textbf{Datasets.}
Main experiments (\cref{main_table}) are conducted on MoNuSeg \cite{kumar2020monuseg, kumar2017monuseg} and TNBC \cite{naylor2019tnbc} datasets. We demonstrate the scalability and robustness of our proposed method (\cref{table_ModelAgnostic}) on four datasets: BRCA \cite{dataset2021brca}, CPM-17 \cite{vu2018cpm}, CryoNuSeg \cite{dataset2021cryonuseg}, and PanNuke \cite{gamper2020pannuke} (see \cref{appendix_dataset} for more details).

\vspace{-0.5cm}

\paragraph{\textbf{Implementation details.}}
Our main experiments use publicly available unsupervised cell instance segmentation (UCIS) models (\emph{i.e.,} SSA \cite{ssa2020miccai} and PSM \cite{psm2023miccai}) as the baseline. Other UCIS models \cite{carpenter2006cellprofiler, schindelin2012fiji, hou2019ieee} are not used since they did not have publicly accessible code. Additionally, for a fair comparison, we follow the standard segmentation evaluation procedures \cite{ssa2020miccai}, employing multi-scale inference and CRF \cite{krahenbuhl2012crf}. Experimentation shows that all USS backbones \cite{caron2021emerging, oquab2023dinov2, he2022masked} performed well, indicating that the choice of backbone had little impact on the results. We conduct all experiments on a single NVIDIA RTX A100 GPU with 80GB memory and implement all USS and UCIS methods in the PyTorch framework.
% In \emph{Step 3}, we maintain the original structure of UCIS baselines, with the additional edge decoder specifically for instance segmentation.
% We leverage Masked Autoencoder (MAE) \cite{he2022masked} as unsupervised semantic segmentation (USS) for all experiments. 
% Comparative experiment of different USS backbones (\emph{i.e.,} DINOv1 \cite{caron2021emerging}, DINOv2 \cite{oquab2023dinov2}, MAE \cite{he2022masked}) is discussed in \cref{appendix_uss}

\vspace{-0.5cm}

\paragraph{\textbf{Evaluation metrics.}}
For a fair comparison with existing cell segmentation approaches, we follow the same evaluation metrics for our results. Specifically, for evaluating cell semantic segmentation performances, we use the Jaccard Index, also known as Intersection over Union (IoU), and Dice \cite{graham2019hovernet}. We assess cell instance segmentation performance through Aggregated Jaccard Index (AJI) \cite{kumar2017aji} and Panotic Quality (PQ) \cite{kirillov2019pq}. FP and FN denote the rate of false positives and false negatives, respectively. \%p indicates percentage point.

\vspace{-0.15cm}

\subsection{\label{mainresults}Comparison with State-of-the-art Approaches}
We benchmark our method against previously proposed state-of-the-art WCIS and UCIS models, reproducing SSA \cite{ssa2020miccai}, PSM \cite{psm2023miccai}, SPPNet \cite{xu2023sppnet}, and TextDiff \cite{feng2024textdiff} under identical conditions. As shown in \cref{main_table}, our method combined with either SSA or PSM consistently outperforms existing approaches, including those that require image-related annotations. Specifically, the cell semantic segmentation performance of SSA improves by at least $+$15\%p, with cell instance segmentation performance achieving a striking twofold enhancement. Consistently, COIN combined with PSM accomplishes at least an $+$8.8\%p in all metrics. 

% 241108 updated
\begin{table*}[t]
    \centering
    \caption{Performance comparison on BRCA \cite{dataset2021brca}, CPM-17 \cite{vu2018cpm}, CryoNuSeg \cite{dataset2021cryonuseg}, and PanNuke \cite{gamper2020pannuke} test datasets.}
    \vspace{-0.3cm}
    \label{table_ModelAgnostic}
    \scriptsize
    \setlength{\tabcolsep}{2pt}
    \renewcommand{\arraystretch}{0.9}
    \begin{threeparttable}
    \begin{tabular}{p{3cm}cccc|cccc|cccc|cccc}
    \toprule
    \textbf{Method} & \multicolumn{4}{c|}{\textbf{BRCA}} & \multicolumn{4}{c|}{\textbf{CPM-17}} & \multicolumn{4}{c|}{\textbf{CryoNuSeg}} & \multicolumn{4}{c}{\textbf{PanNuke}} \\
    \cmidrule(lr){2-5} \cmidrule(lr){6-9} \cmidrule(lr){10-13} \cmidrule(lr){14-17}
     & \textbf{AJI} ($\uparrow$)& \textbf{PQ} ($\uparrow$)& \textbf{IoU} ($\uparrow$)& \textbf{Dice} ($\uparrow$)& \textbf{AJI} ($\uparrow$)& \textbf{PQ} ($\uparrow$)& \textbf{IoU} ($\uparrow$)& \textbf{Dice} ($\uparrow$)& \textbf{AJI} ($\uparrow$)& \textbf{PQ} ($\uparrow$)& \textbf{IoU} ($\uparrow$)& \textbf{Dice} ($\uparrow$)& \textbf{AJI} ($\uparrow$)& \textbf{PQ} ($\uparrow$)& \textbf{IoU} ($\uparrow$)& \textbf{Dice} ($\uparrow$)\\
    \midrule
    SSA \cite{ssa2020miccai} {\tiny MICCAI'20} & 0.205 & 0.136 & 0.590 & 0.508 & 0.444 & 0.378 & 0.735 & 0.734 & 0.212 & 0.145 & 0.637 & 0.641 & 0.395 & 0.245 & 0.683 & 0.671 \\
    \rowcolor{gray!20}\textbf{SSA + COIN (Ours)} & \textbf{0.298} & \textbf{0.216} & \textbf{0.620} & \textbf{0.546} & \textbf{0.610} & \textbf{0.532} & \textbf{0.810} & \textbf{0.821} & \textbf{0.307} & \textbf{0.296} & \textbf{0.717} & \textbf{0.741} & \textbf{0.478} & \textbf{0.400} & \textbf{0.741} & \textbf{0.743} \\
    \rowcolor{gray!20}$\Delta_{ssa}$ & \textcolor{blue}{+0.093} & \textcolor{blue}{+0.080} & \textcolor{blue}{+0.030} & \textcolor{blue}{+0.038} & \textcolor{blue}{+0.166} & \textcolor{blue}{+0.154} & \textcolor{blue}{+0.075} & \textcolor{blue}{+0.087} & \textcolor{blue}{+0.095} & \textcolor{blue}{+0.151} & \textcolor{blue}{+0.080} & \textcolor{blue}{+0.100} & \textcolor{blue}{+0.083} & \textcolor{blue}{+0.155} & \textcolor{blue}{+0.058} & \textcolor{blue}{+0.072} \\
    \bottomrule
    \end{tabular}
    \begin{tablenotes}
        \item Blue indicates a favorable change in performance.
        \item $\Delta_{ssa}$: Performance gap between SSA \cite{ssa2020miccai} and our proposed method.
    \end{tablenotes}
    \end{threeparttable}
    \vspace{-0.3cm}
\end{table*}

\begin{table}[t]
    \centering
    \caption{Comparison of two UCIS methods \cite{ssa2020miccai, psm2023miccai} on MoNuSeg \cite{kumar2017monuseg, kumar2020monuseg} train and test sets.}
    \vspace{-0.3cm}
    \label{traintest_ablation}
    \scriptsize
    \setlength{\tabcolsep}{1.5pt} 
    \renewcommand{\arraystretch}{0.9} 
    \begin{threeparttable}
    \begin{tabular}{lcccc|cccc}
        \toprule
        \textbf{Method} & \multicolumn{4}{c|}{\textbf{Train}} & \multicolumn{4}{c}{\textbf{Test}} \\
        \cmidrule(lr){2-5} \cmidrule(lr){6-9}
         & \textbf{AJI} ($\uparrow$)& \textbf{IoU} ($\uparrow$)& \textbf{FN} ($\downarrow$)& \textbf{FP} ($\downarrow$)& \textbf{AJI} ($\uparrow$)& \textbf{IoU} ($\uparrow$)& \textbf{FN} ($\downarrow$)& \textbf{FP} ($\downarrow$)\\
        \midrule
        SSA \cite{ssa2020miccai} {\tiny MICCAI'20} & 0.190 & 0.540 & 0.289 & 0.171 & 0.259 & 0.618 & 0.234 & 0.149 \\
        \rowcolor{gray!20} \textbf{SSA + COIN} & \textbf{0.445} & \textbf{0.716} & \textbf{0.160} & \textbf{0.124} & \textbf{0.580} & \textbf{0.776} & \textbf{0.116} & \textbf{0.107} \\
        \rowcolor{gray!20}$\Delta_{ssa}$ & \textcolor{blue}{+0.255} & \textcolor{blue}{+0.176} & \textcolor{blue}{-0.129} & \textcolor{blue}{-0.047} & \textcolor{blue}{+0.321} & \textcolor{blue}{+0.158} & \textcolor{blue}{-0.118} & \textcolor{blue}{-0.042} \\
        \midrule
        PSM \cite{psm2023miccai} {\tiny MICCAI'23} & 0.344 & 0.607 & 0.250 & 0.143 & 0.471 & 0.690 & 0.187 & 0.124 \\
        \rowcolor{gray!20} \textbf{PSM + COIN} & \textbf{0.463} & \textbf{0.733} & \textbf{0.147} & \textbf{0.120} & \textbf{0.579} & \textbf{0.777} & \textbf{0.112} & \textbf{0.111} \\
        \rowcolor{gray!20}$\Delta_{psm}$ & \textcolor{blue}{+0.119} & \textcolor{blue}{+0.126} & \textcolor{blue}{-0.103} & \textcolor{blue}{-0.023} & \textcolor{blue}{+0.108} & \textcolor{blue}{+0.087} & \textcolor{blue}{-0.075} & \textcolor{blue}{-0.013} \\
        \bottomrule
    \end{tabular}
    \begin{tablenotes}
        \item Blue indicates a favorable change in performance.
        \item $\Delta_{ssa}$: Performance gap between SSA \cite{ssa2020miccai} and our proposed method.
        \item $\Delta_{psm}$: Performance gap between PSM \cite{psm2023miccai} and our proposed method.
    \end{tablenotes}
    \end{threeparttable}
    \vspace{-0.3cm}
\end{table}

\vspace{-0.1cm}
\subsection{\label{discussion}Discussion}
\vspace{-0.1cm}
\paragraph{\textbf{Scalability.}} \cref{table_ModelAgnostic} highlights that our COIN consistently improves the UCIS baseline (\emph{e.g.}, SSA \cite{ssa2020miccai}) across six benchmarks \cite{naylor2019tnbc, dataset2021brca, dataset2021cryonuseg, gamper2020pannuke, kumar2017monuseg, kumar2020monuseg}. As depicted in \cref{appendix_ModelAgnostic}, our approach improves SSA \cite {ssa2020miccai} by significantly reducing FN and FP.
\vspace{-0.5cm}
\paragraph{\textbf{Flexibility.}} 
To evaluate the quality of pseudo labels, we examine COIN’s performance on the MoNuSeg train and test sets. As shown in \cref{traintest_ablation}, integrating COIN with existing UCIS models, such as SSA \cite{ssa2020miccai} and PSM \cite{psm2023miccai}, consistently improves performance across both sets. Specifically, integrating SSA with our COIN yields a substantial increase in AJI by $+$25.5\%p on the train set and $+$32.1\%p on the test set. Similarly, combining our method with PSM results in AJI score enhancements of $+$11.9\%p on the train set and $+$10.8\%p on the test set. Moreover, on the train set, FN and FP significantly decrease by $-$0.129 and $-$0.047 for SSA, and by $-$0.103 and $-$0.023 for PSM. %respectively, highlighting the model-agnostic nature of COIN. See \Cref{appendix_modelagnostic_figure} for qualitative results.
\vspace{-0.5cm}
\paragraph{\textbf{Effect of pixel-level cell propagation.}}
COIN combines unsupervised semantic segmentation (USS) \cite{he2022masked} with optimal transport (OT) \cite{rachev1985monge}, significantly improving instance sensitivity and generating error-free instances (\cref{problem}). As shown in \cref{propagation_ablation}, USS propagation alone (second row) reduces FN by $-$0.516, indicating better detection of instances. However, the imprecise cell boundaries result in only a marginal increase in the top 5\% AJI score and a substantial rise in FP ($+$0.423). To address this, we incorporate OT into the propagation process (third row), which refines edge boundaries by focusing on the minor class (\emph{e.g.,} cells with fewer pixels). Therefore, as previously shown in \cref{OT_effect}, the cell and tissue regions become clearly distinguished, showing a significant reduction in FP by $-$0.249. 

%achieving instances with a near-perfect top 5\% AJI score (\emph{i.e.,} close to 1), which are considered as error-free instances (see \cref{problem}).} 

%A comparative experiment on different USS backbones (\emph{i.e.,} DINOv1 \cite{caron2021emerging}, DINOv2 \cite{oquab2023dinov2}, MAE \cite{he2022masked}) is discussed in \cref{appendix_uss}, where MAE demonstrated the best performance and was therefore selected for all experiments.

\paragraph{\textbf{USS and OT alternatives.}}
In \cref{appendix_uss}, the experiment with various USS alternatives \cite{caron2021emerging, oquab2023dinov2, he2022masked} showed MAE \cite{he2022masked} with marginal advantage, so we used MAE for all experiments. After fixing USS, comparisons between various clustering methods indicated that OT was the fastest and most effective approach for elevated FP. As demonstrated in \cref{ot_alternatives}, compared to CRF \cite{krahenbuhl2012crf}, K-means clustering (K-means) \cite{macqueen1967kmeans}, and Gaussian mixture models (GMM) \cite{neal1998gmm}, OT outperforms all of them by at least $+$5.7\%p increase in IoU and  1.8$\times$ reduction in FP, while being up to 50$\times$ faster in computational cost. % appendix에서 theoretical superiority 설명
\vspace{-0.4cm}

\begin{table}[t]
    \centering
    % \caption{Effect of pixel-level cell propagation combined with Optimal Transport \cite{rachev1985monge} on the MoNuSeg \cite{kumar2017monuseg, kumar2020monuseg} test set (\emph{Step 1} feature ablation).}
    \caption{Effect of pixel-level cell propagation (\emph{Step 1}) combined with Optimal Transport \cite{rachev1985monge} on the MoNuSeg \cite{kumar2017monuseg, kumar2020monuseg} test set.}
    \vspace{-0.3cm}
    \label{propagation_ablation}
    \scriptsize
    \setlength{\tabcolsep}{2pt} 
    \renewcommand{\arraystretch}{0.9} 
    \begin{threeparttable}
    \begin{tabular}{ccccccc}
        \toprule
        \textbf{USS} & {\textbf{OT}} & {\textbf{Instance-level}} & \multicolumn{3}{c}{\textbf{Pixel-level}} \\
        \cmidrule(lr){3-3} \cmidrule(lr){4-7}
         \textbf{\cref{eq_S_us}} & \textbf{\cref{eq_S_OT}} & \textbf{AJI (Top 5\%)} ($\uparrow$)& \textbf{IoU} ($\uparrow$)& \textbf{FN} ($\downarrow$)& \textbf{FP} ($\downarrow$)& \\
        \midrule
        \xmark& \xmark& 0.734 & 0.305 & 0.536 & 0.159\\
        \cmark& \xmark& 0.739 (\textcolor{blue}{+0.005})& 0.439 (\textcolor{blue}{+0.134}) & 0.020 (\textcolor{blue}{-0.516})& 0.552 (\textcolor{red}{+0.423})\\
        \rowcolor{gray!20} \cmark & \cmark & \textbf{0.985} (\textcolor{blue}{+0.251}) & \textbf{0.543} (\textcolor{blue}{+0.238}) & \textbf{0.157} (\textcolor{blue}{-0.379})& \textbf{0.301} (\textcolor{red}{+0.142})\\
        \bottomrule
    \end{tabular}
    \end{threeparttable}
    \begin{tablenotes}
        \item Blue indicates a favorable change in performance, and red indicates an unfavorable change in performance.
    \end{tablenotes}
    \vspace{-0.2cm}
\end{table}

% \begin{table}[t]
%     \centering
%     \caption{Effect of USS propagation combined with Optimal Transport \cite{rachev1985monge} on the MoNuSeg \cite{kumar2017monuseg, kumar2020monuseg} test set.}
%     \vspace{-0.3cm}
%     \label{propagation_ablation}
%     \scriptsize
%     \setlength{\tabcolsep}{0.25pt} 
%     \renewcommand{\arraystretch}{1} 
%     \begin{threeparttable}
%     \begin{tabular}{cccccccc}
%         \toprule
%         \textbf{USS} & {\textbf{OT}} & \textbf{Watershed} & {\textbf{Instance-level}} & \multicolumn{3}{c}{\textbf{Pixel-level}} \\
%         \cmidrule(lr){4-4} \cmidrule(lr){5-8}
%          \textbf{\cref{eq_S_us}} & \textbf{\cref{eq_S_OT}} & \textbf{\cref{step2}} & \textbf{AJI (Top 5\%)} ($\uparrow$)& \textbf{IoU} ($\uparrow$)& \textbf{FN} ($\downarrow$)& \textbf{FP} ($\downarrow$)& \\
%         \midrule
%         \xmark & \xmark & \xmark& 0.734 & 0.305 & 0.536 & 0.159\\
%         \cmark & \xmark  & \xmark & 0.739 (\textcolor{blue}{+0.005})& 0.439 (\textcolor{blue}{+0.134}) & 0.020 (\textcolor{blue}{-0.516})& 0.552 (\textcolor{red}{+0.423})\\
%         \cmark & \cmark  & \xmark & ? (\textcolor{blue}{+0.000}) & 0.539 (\textcolor{blue}{+0.234}) & 0.163 (\textcolor{blue}{-0.373})& 0.303 (\textcolor{red}{+0.144})\\
%         \rowcolor{gray!20} \cmark & \cmark & \cmark & 0.985 (\textcolor{blue}{+0.251}) & 0.543 (\textcolor{blue}{+0.238}) & 0.157 (\textcolor{blue}{-0.379})& 0.301 (\textcolor{red}{+0.142})\\
%         \bottomrule
%     \end{tabular}
%     \end{threeparttable}
%     \vspace{-0.2cm}
% \end{table}

\begin{table}[t]
    \centering
    % \caption{Component-wise complexity and OT alternatives evaluation.}
    \caption{Performance and complexity analysis of OT \cite{rachev1985monge}.}
    \vspace{-0.3cm}
    \scriptsize
    \setlength{\tabcolsep}{6pt}
    \renewcommand{\arraystretch}{0.9}
    \begin{tabular}{lccc}
        \toprule
        \textbf{Component} & \textbf{Latency per Image} & \textbf{IoU ($\uparrow$)} & \textbf{FP($\downarrow$)}\\
        \midrule
        USS (\emph{i.e.,} MAE \cite{he2022masked})    & 330 ms  & 0.439 & 0.552\\
        USS + CRF \cite{krahenbuhl2012crf} & 330ms + 200 ms  & 0.443 \textcolor{blue}{(+0.004)} & 0.549 \textcolor{blue}{(-0.003)}\\
        USS + K-means \cite{macqueen1967kmeans} & 330ms + 540 ms  & 0.464 \textcolor{blue}{(+0.025)} & 0.531 \textcolor{blue}{(-0.021)} \\
        % USS + GMM \cite{neal1998gmm} & 330ms + 580 ms  & \underline{0.482} \textcolor{blue}{(+0.043)} & \underline{0.484} \textcolor{blue}{(-0.068)}\\
        USS + GMM \cite{neal1998gmm} & 330ms + 580 ms  & {0.482} \textcolor{blue}{(+0.043)} & {0.484} \textcolor{blue}{(-0.068)}\\
        \rowcolor{gray!20} USS + OT \cite{rachev1985monge} & 330ms + 10 ms   & \textbf{0.539} \textcolor{blue}{(+0.100)} & \textbf{0.303} \textcolor{blue}{(-0.249)}\\
        \bottomrule
    \end{tabular}
    \begin{tablenotes}
        \item Blue indicates a favorable change in performance.
    \end{tablenotes}
    \label{ot_alternatives}
    \vspace{-0.2cm}
\end{table}

\begin{table}[t]
    \centering
    \caption{Effect of self-distillation with instance-level confidence scoring on MoNuSeg \cite{kumar2017monuseg, kumar2020monuseg} train and test sets (\emph{Step 1} vs. \emph{Step 2\&3} ablation).}
    \vspace{-0.3cm}
    \label{recursive_ablation}
    \scriptsize
    \renewcommand{\arraystretch}{0.9} 
    \begin{threeparttable}
    \begin{tabular}{cccccc}
        \toprule
        \textbf{Instance} & {\textbf{Distillation}} & &&&\\
        \textbf{Propagation} & {\textbf{w/ Scoring}} & \multicolumn{2}{c}{\textbf{Train}} & \multicolumn{2}{c}{\textbf{Test}} \\
        \cmidrule(lr){3-4} \cmidrule(lr){5-6}
         \textbf{\cref{step1}} & \textbf{\cref{step2} \& \ref{step3}} & \textbf{AJI} ($\uparrow$)& \textbf{IoU} ($\uparrow$)& \textbf{AJI} ($\uparrow$)& \textbf{IoU} ($\uparrow$) \\
        \midrule
        \xmark & \xmark & 0.190 & 0.540 & 0.259 & 0.618 \\
        \cmark & \xmark & 0.233 & 0.537 & 0.260 & 0.559 \\
        \xmark & \cmark & 0.331 & 0.665 & 0.386 & 0.678 \\
        \rowcolor{gray!20} 
        % \cmark & \cmark & 0.445 (\textcolor{blue}{+0.255}) & 0.716 (\textcolor{blue}{+0.176}) & 0.580 (\textcolor{blue}{+0.321}) & 0.776 (\textcolor{blue}{+0.158}) \\
        \cmark & \cmark & \textbf{0.445} & \textbf{0.716} & \textbf{0.580} & \textbf{0.776} \\
      \bottomrule
    \end{tabular}
    \end{threeparttable}
    \vspace{-0.4cm}
\end{table}
\paragraph{\textbf{Effect of self-distillation with scoring.}} 
We evaluate the impact of instance-level confidence scoring on recursive self-distillation in \cref{recursive_ablation}. Performance metrics are assessed on the MoNuSeg train and test sets, which reveal consistent improvement patterns. Compared to the baseline model (the first row), employing just the instance-level confidence scoring without pixel-level cell propagation (the third row) results in performance enhancement with $+$12.5\%p in IoU and $+$14.1\%p in AJI for train set and  $+$6\%p in IoU and $+$12.7\%p in AJI for test set. Overall, our complete model (the fourth row) compared to the baseline shows significant performance improvement with at least $+$15.8\%p in IoU of train and test sets. Likewise, the AJI score doubles on train and test sets for cell instance segmentation performance. %demonstrating a substantial improvement in cell instance segmentation.

\vspace{-0.5cm}

% \paragraph{\textbf{Application of SAM for unsupervised scoring.}}
\paragraph{\textbf{Addressing SAM's sensitivity in unsupervised scoring.}}
% We examine the rationale behind SAM's application in \cref{sam_ablation}. 
When SAM is applied without our instance-level scoring (the second row in \cref{sam_ablation}), all metrics drop to zero on the test set. This failure arises from SAM's sensitivity to point prompts, specifically when predictions inadvertently target background pixels, as demonstrated in \cref{InstanceScoring}. If only SAM is used, over-propagation causes detection to fail completely. 
% To mitigate this, we use SAM exclusively to validate the model's outputs rather than rely on SAM predictions. 
To mitigate this, we address the issue by selecting only instance masks with high consistency (\emph{i.e.}, IoU) between SAM and the model's outputs.
As shown in \cref{appendix_Scoring_fig}, the propagated masks generated by the UCIS baseline \cite{ssa2020miccai} appear coarse and ambiguous, struggling to distinguish between cells and tissues. Through our scoring method, distinct cell instances are accurately recognized, with each assigned a corresponding confidence score represented by color. By training only on instances with high consistency, COIN effectively reduces the uncertainty in SAM. In short, we propose a novel solution to the prompt sensitivity of SAM without relying on supervision by selectively using SAM predictions as pseudo-GT only for cell-targeted point prompts (see \cref{appendix_sam} for more details).
% we offer a new solution to the prompt sensitivity of SAM with point prompts without depending on supervision (see \cref{appendix_sam} for more details).

\begin{table}[t]
    \centering
    \caption{Effect of our instance scoring compared to the naïve SAM \cite{kirillov2023segment} with SSA \cite{ssa2020miccai} (baseline) on the MoNuSeg \cite{kumar2017monuseg, kumar2020monuseg} test set (\emph{Step 2} feature ablation).}
    \vspace{-0.3cm}
    \label{sam_ablation}
    \scriptsize
    \setlength{\tabcolsep}{1.5pt} 
    \renewcommand{\arraystretch}{0.9} 
    \begin{threeparttable}
    \begin{tabular}{>{\centering}p{0.08\textwidth} cccc}
        \toprule
        \textbf{SAM} & {\textbf{Scoring (\cref{step2})}} & \textbf{AJI} ($\uparrow$)& \textbf{PQ} ($\uparrow$)& \textbf{Dice} ($\uparrow$) \\
        \midrule
        \xmark & \xmark & 0.259 & 0.185 & 0.575 \\
        \cmark & \xmark & 0.000 (\textcolor{red}{-0.259}) & 0.000 (\textcolor{red}{-0.185}) & 0.000 (\textcolor{red}{-0.575}) \\
        \rowcolor{gray!20} 
        \cmark & \cmark & \textbf{0.580} (\textcolor{blue}{+0.321}) & \textbf{0.536} (\textcolor{blue}{+0.351}) & \textbf{0.794} (\textcolor{blue}{+0.219}) \\
      \bottomrule
    \end{tabular}
    \end{threeparttable}
    \begin{tablenotes}
        \item Blue indicates a favorable change in performance, and red indicates an unfavorable change in performance.
    \end{tablenotes}
    \vspace{-0.35cm}
\end{table}

% \begin{figure}[t]
%   \centering
%   \includegraphics[width=0.95\linewidth]{ICCV 2025/figures_main/Figure_SAM.drawio.pdf}
%   \vspace{-0.2cm}
%   \caption{\textbf{Visualization of SAM's \cite{kirillov2023sam} success and failure cases.} SAM's predictions may be erroneous when point prompts target background pixels (bottom), causing most pixels to be misclassified as foreground.}
%   \vspace{-0.2cm}
%   \label{figure_sam}
% \end{figure}

\begin{figure}[t]
  \centering
  \includegraphics[width=0.9\linewidth]{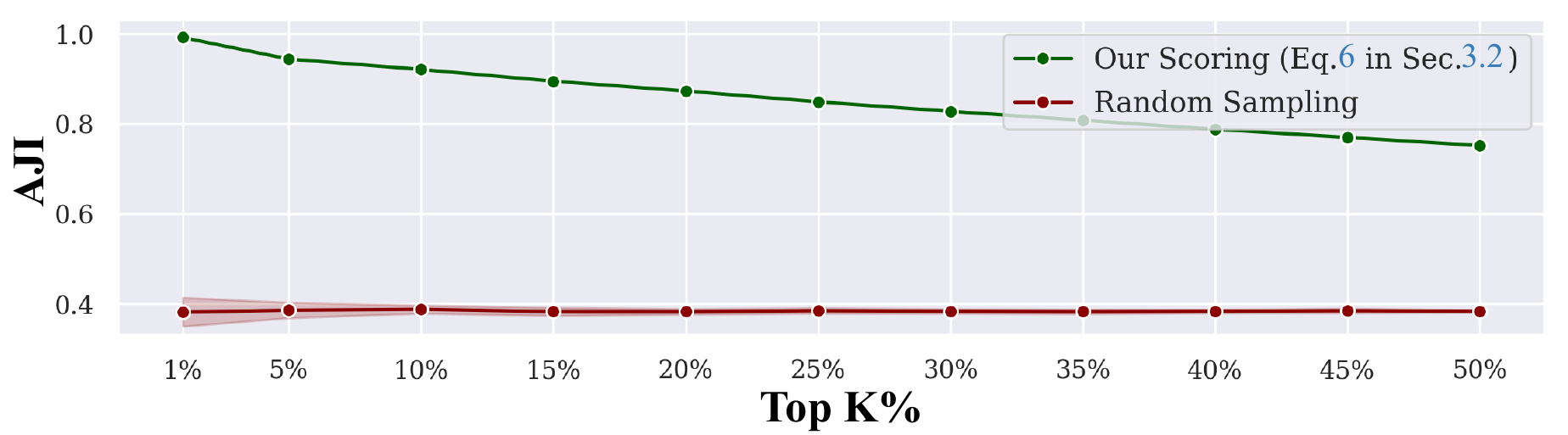}
  \vspace{-0.35cm}
  \caption{\textbf{Effect of our scoring method on the MoNuSeg \cite{kumar2017monuseg, kumar2020monuseg} train set.} Without relying on actual ground truth, our scoring approach yields highly confident instances that closely resemble the true instances (\emph{i.e.}, AJI = 1).}
  \label{figure_correlation}
  \vspace{-0.6cm}
\end{figure}

As shown in \cref{figure_correlation}, we further compare our scoring with random sampling to validate the effectiveness of consistency comparison in identifying error-free instances. The x-axis and y-axis indicate the top K\% of instances and the corresponding AJI scores, respectively. The results show that our instance-level scoring, even without instance annotations, successfully identifies highly confident instances, whereas random sampling does not. For the top 1\% of instances, our approach achieves an AJI close to 1 (\emph{i.e.,} error-free), significantly outperforming random sampling that fails to exceed an AJI of 0.5.

\vspace{-0.4cm}

\paragraph{\textbf{Hyperparameters.}} 
For a fair comparison, we strictly adopt watershed \cite{ssa2020miccai} and CRF \cite{krahenbuhl2012crf} parameters from UCIS baselines \cite{ssa2020miccai, psm2023miccai}, with $\lambda$ of OT in \cref{eq_S_OT} fixed at 0.1. The experiment shown in \cref{appendix_ot_lambda} confirmed that 0.1 is the best value, with consistent IoU scores in the range of 0.01 to 0.45, demonstrating the stability of COIN in $\lambda$. No parameters are required for USS \cite{he2022masked} and SAM \cite{kirillov2023segment}. 

\vspace{-0.4cm}

\paragraph{Computational complexity.} \label{comp_copmplexity}
In our framework, measuring instance-level consistency with SAM takes 9.5 milliseconds per instance. To acquire high-resolution USS feature maps, we partition each image into six equal, non-overlapping patches horizontally and vertically, which takes 130 milliseconds per patch for processing. As demonstrated in \cref{rebutt_comp_complexity}, the inference time remains the same for both MoNuSeg \cite{kumar2017monuseg, kumar2020monuseg} and TNBC \cite{naylor2019tnbc} datasets since all components (\emph{e.g.,} USS) are used only during training. In \cref{step3}, an edge decoder was introduced to the baseline UCIS \cite{ssa2020miccai} to strengthen instance edge separation. Despite this addition, the overall inference time remained nearly unchanged, with only 1-3 ms added per image. While the 50\% increase in training time is a limitation, COIN achieves more than a twofold improvement in AJI over existing UCIS methods \cite{ssa2020miccai, psm2023miccai} and, for the first time, surpasses semi- and weakly-supervised models (\emph{e.g.}, using points) without additional annotations (\cref{main_table} and \cref{traintest_ablation}).
\vspace{-0.2cm}
% While the 50\% increase in training time is a limitation, COIN improves AJI by twofold for MoNuSeg and surpasses the performance of weakly supervised models (\emph{e.g.}, using points) (\cref{main_table} and \cref{traintest_ablation}).

\begin{table}[t]
    \centering
    \caption{Analysis of computational complexity.}
    \label{rebutt_comp_complexity}
    \vspace{-0.3cm}
    \scriptsize
    \setlength{\tabcolsep}{3pt}
    \renewcommand{\arraystretch}{0.9}
    \begin{threeparttable}
    \begin{tabular}{lcc}
        \toprule
        \textbf{Phase \textcolor{purple}{(Dataset)}} & \textbf{SSA \cite{ssa2020miccai} w/o COIN} & \textbf{SSA \cite{ssa2020miccai} w/ COIN} \\
        \midrule
        Total Training Time \textcolor{purple}{(MoNuSeg)} & 8 hours & 14 hours \textcolor{red}{(+6 hours)} \\
        Total Testing Time \textcolor{purple}{(MoNuSeg)}  & 1 minute & 1 minute \textcolor{blue}{(+0 minutes)} \\
        \midrule
        Total Training Time \textcolor{purple}{(TNBC)} & 4 hours & 6 hours \textcolor{red}{(+2 hours)} \\
        Total Testing Time \textcolor{purple}{(TNBC)}  & 1 minute & 1 minute \textcolor{blue}{(+0 minutes)} \\
        \bottomrule
    \end{tabular}
    \end{threeparttable}
    \begin{tablenotes}
        \item Blue indicates a favorable change in performance, and red indicates an unfavorable change in performance.
    \end{tablenotes}
    \vspace{-0.4cm}
\end{table}

\section{Conclusion}
\vspace{-0.2cm}
In this paper, we present COIN, a three-step approach that overcomes \emph{the absence of error-free instances} (\cref{problem}), inherent in existing UCIS methods (\emph{i.e.,} SSA \cite{ssa2020miccai}, PSM \cite{psm2023miccai}), through instance-level confidence scoring approach combined with recursive self-distillation (\cref{overview}). Notably, our method achieves substantial performance improvement in instance segmentation by more than twofold in SSA and at least $+$18\%p in PSM on MoNuSeg (\cref{main_table}). Our extensive experiments demonstrate that while COIN operates without relying on any image-related annotations, it consistently outperforms supervised models \cite{xu2023sppnet,feng2024textdiff}. We believe that our simple yet powerful approach will provide valuable insight into future research on cell instance segmentation, as our instance-level confidence scoring offers a new perspective on uncertainty estimation in cell segmentation tasks. Moreover, COIN sets the foundation for developing more robust medical image analysis tools that do not require image-related annotations, thereby supporting faster and more accurate clinical decision-making.

\section*{Acknowledgments}
{\small Kyungsu Kim is affiliated with the School of Transdisciplinary Innovations, Department of Biomedical Science, Medical Research Center, Interdisciplinary Program in Bioengineering, and Interdisciplinary Program in Artificial Intelligence at Seoul National University, Seoul, Republic of Korea. Hyungseok Seo and Seo Jin Lee are affiliated with the Laboratory of Cell \& Gene Therapy, Institute of Pharmaceutical Sciences, College of Pharmacy, Seoul National University, Korea. Sanghyun Jo and Seungwoo Lee are affiliated with OGQ GYN, Seoul, Korea. Seohyung Hong is affiliated with the Department of Biomedical Science and Medical Research Center, Seoul National University, Republic of Korea.}

{\small This work was supported by the Institute of Information \& Communications Technology Planning \& Evaluation (IITP) grant funded by the Korea government (MSIT) [RS-2025-02305581], [RS-2025-25442338 (AI Star Fellowship Support Program at SNU)], and [RS-2021-II211343 (Artificial Intelligence Graduate School Program at SNU)]. This work was supported by grants of the MD–PhD/Medical Scientist Training Program and Korea Health Technology R\&D Project through the Korea Health Industry Development Institute (KHIDI), funded by the Ministry of Health \& Welfare, Republic of Korea [RS-2025-02307233], and the National Research Foundation of Korea (NRF) grants (RS-2023-00242443, RS-2023-00282907), both funded by the Korean government (MSIT). This research was results of a study on the AI Media and Cultural Enjoyment Expansion Project, supported by the Ministry of Science and ICT and NIPA in 2025. H.S. was funded through the Creative-Pioneering Researchers Program at Seoul National University.}
% {\small This work was supported by a grant from the Institute of Information \& Communications Technology Planning \& Evaluation (IITP) [RS-2025-02305581], a grant from the IITP [RS2021-II211343, Artificial Intelligence Graduate School Program, Seoul National University], a grant from the Korea Health Technology R\&D Project through the Korea Health Industry Development Institute (KHIDI), funded by the Ministry of Health \& Welfare, Republic of Korea (grant number : RS-2025-02307233), and the National Research Foundation of Korea (NRF) grant (RS-2023-00242443, RS-2023-00282907), funded by the Korean government (MSIT). This research was conducted as part of the "AI Media and Cultural Enjoyment Expansion" Project, supported by the Ministry of Science and ICT and the National IT Industry Promotion Agency (NIPA) in 2025.}

{
    \small
    \bibliographystyle{ieeenat_fullname}
    \bibliography{main}
}

\clearpage
\setcounter{page}{1}

\appendix
\renewcommand{\thesection}{\Alph{section}}

\footnotetext[1]{These authors contributed equally.}
\footnotetext[2]{Corresponding author.}

\begin{strip}
  \begin{center}
    \Large{Supplementary Material} \\
    \vspace{0.2cm}
    \large{COIN: Confidence Score-Guided Distillation for Annotation-Free Cell Segmentation} \\
    \vspace{-0.1cm}
  \end{center}
%\end{strip}

% ── Authors (full–width) ───────────────────────────────
%\begin{strip}
  %\vspace{-2cm}
  \begin{center}\normalsize
    Sanghyun Jo$^{1}$\footnotemark[1]\quad
    Seo Jin Lee$^{3}$\footnotemark[1]\quad
    Seungwoo Lee$^{1}$\quad
    Seohyung Hong$^{4}$\\[0.5ex]
    Hyungseok Seo$^{3}$\footnotemark[2]\quad
    Kyungsu Kim$^{2,4,5}$\footnotemark[2]\\[0.5ex]
    \texttt{%
      \{shjo.april, vict.lee0\}@gmail.com \quad
      \{seojinleee, hong.sh, h.seo, kyskim\}@snu.ac.kr%
    }
  \end{center}
%\end{strip}

% ── Affiliations (full–width) ──────────────────────────
%\begin{strip}
  \begin{center}\small
    \vspace{-0.2cm}
    $^{1}$OGQ, Seoul, Korea
    \quad $^{2}$School of Transdisciplinary Innovations, Seoul National University, Korea\\
    $^{3}$Laboratory of Cell \& Gene Therapy, Institute of Pharmaceutical Sciences, College of Pharmacy, Seoul National University, Korea\\
    $^{4}$Department of Biomedical Science and Medical Research Center, College of Medicine, Seoul National University, Korea\\
    $^{5}$Interdisciplinary Programs in Artificial Intelligence, Bioengineering, and Bioinformatics, Seoul National University, Korea
  \end{center}
%\end{strip}

%\begin{strip}
\centering
\includegraphics[width=0.90\textwidth]{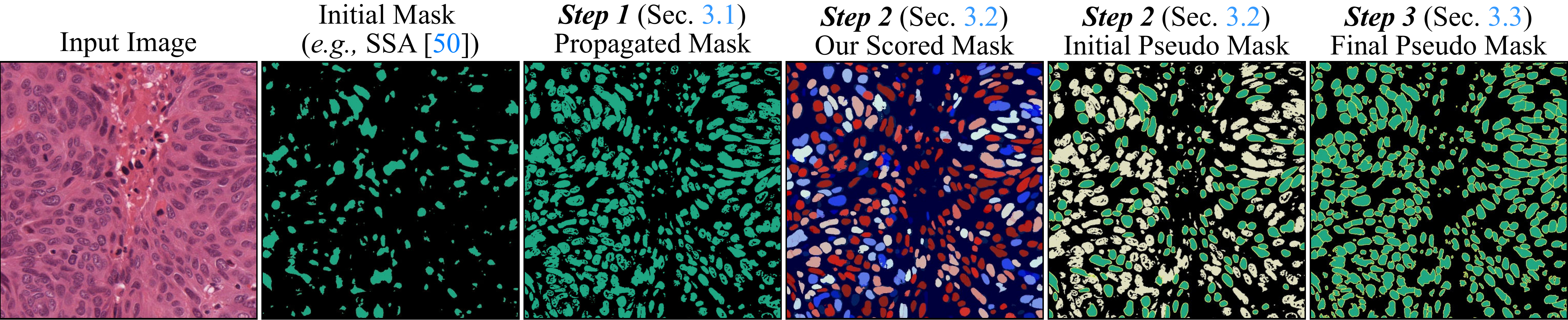}
\vspace{-0.35cm}
\captionof{figure}{A brief overview of three steps in COIN.}
\label{appendix_overview}
\end{strip}

% \begin{figure}[t]
%   \centering
%   \includegraphics[width=0.9\linewidth]{ICCV 2025/figures_main/Figure_Correlation.pdf}
%   \vspace{-0.35cm}
%   \caption{\textbf{Effect of our scoring method on the MoNuSeg \cite{kumar2017monuseg, kumar2020monuseg} train set.} Without relying on actual ground truth, our scoring approach yields highly confident instances that closely resemble the true instances (\emph{i.e.}, AJI = 1).}
%   \label{figure_correlation}
%   \vspace{-0.6cm}
% \end{figure}

\section{Method Overview}
Our approach is divided into three steps, and we assess the effect of each step in \cref{appendix_overview_table} on adjacent and non-adjacent cells (see \cref{appendix_adjacentcell_fig} for more information). In \emph{Step 1}, pixel-level cell propagation utilizing USS \cite{he2022masked} was used to increase the sensitivity to detect all instances, resulting in a significant drop in false negative rate (FN). Specifically, FN decreases 2.3-fold, corresponding to \cref{appendix_overview} that illustrates how most cells become detected following \emph{Step 1} (the second row). However, this was accompanied by an increase in the rate of false positives (FP), which was handled by incorporating optimal transport (OT) \cite{rachev1985monge} for its ability to cluster minor pixel groups. In \emph{Step 2}, to identify and use only error-free instances for recursive self-distillation, we introduce, for the first time, an instance-level confidence scoring approach to automatically select highly confident instances without depending on the ground truth (GT) annotations. This scoring approach measures the consistency between the baseline UCIS model \cite{ssa2020miccai} prediction and SAM-generated mask and selects only the instances close to GT (\emph{i.e.,} instances with AJI scores close to 1). As shown in the table (the third row) significantly decreases FP, particularly threefold for non-adjacent cells. {Here, consistency-based selection acts as implicit memory that preserves error-free masks with the dynamically adjusted threshold $\delta_k$ (\cref{eq_decision}), preventing the accumulation of noisy labels. Additionally, we chose to decouple pseudo-label generation from training (USS once/image, SAM each epoch) to avoid end-to-end fine-tuning yet still double AJI without increasing the inference time (Tabs. \ref{main_table}, \ref{traintest_ablation}, and \ref{rebutt_comp_complexity}). Training slows by 50-75\% compared to the baseline \cite{ssa2020miccai} (\cref{appendix_overview_table}), and our modular design supports multiple UCIS (SSA \cite{ssa2020miccai}, PSM \cite{psm2023miccai}; Tabs. \ref{main_table} and \ref{traintest_ablation}) and USS models  \cite{caron2021emerging, oquab2023dinov2, darcet2023vision, he2022masked} (\cref{appendix_uss_table}).} Then, in \emph{Step 3}, the selected instances are used for recursive self-distillation to expand the confidence, progressively increasing the number of highly confident instances each round. Notably, this last step results in a 1.9-fold improvement in AJI for adjacent cells (the fourth row), highlighting substantial advancements in our method's accuracy in cell instance segmentation, as depicted in \cref{appendix_overview}.
%Our method is divided into three steps, and we assess the effect of each step \cref{appendix_overview_table}. Beginning with \emph{Step 1} (\cref{step1}), we focus on minimizing false negatives (FN) by incorporating optimal transport (OT) \cite{rachev1985monge}, but it still yields high false positives (FP) (the second row). Specifically, FN decreases by 2.3-fold, corresponding to \cref{appendix_overview} that illustrates how most cells become detected following \emph{Step 1}. To address the accompanying increase in FP, we introduce instance-level confidence scoring to automatically select confident instances highly consistent with SAM-based predictions. This scoring approach significantly decreases FP, particularly threefold for non-adjacent cells (the third row). Finally, by applying self-distillation in \emph{Step 3}, we minimize both FN and FP and significantly enhance segmentation performance (the fourth row). Notably, this last step results in a 1.9-fold improvement in AJI for adjacent cells, highlighting substantial advancements in our method's accuracy in cell instance segmentation, as depicted in \cref{appendix_overview}.

\begin{table}[t]
    \setlength{\tabcolsep}{8pt}
    \renewcommand{\arraystretch}{1.2} 
    \caption{Performance comparison of the three steps of COIN on adjacent and non-adjacent cells on the MoNuSeg \cite{kumar2017monuseg, kumar2020monuseg} train set.}
    \label{appendix_overview_table}
    \vspace{-0.35cm}
    \scriptsize
    \centering
    \setlength{\tabcolsep}{4pt}
    \renewcommand{\arraystretch}{1.0}
    \begin{threeparttable}
        \begin{tabular}{lcccc|cccc} 
            \toprule
            {\textbf{Method}} & \multicolumn{4}{c}{\textbf{Non-adjacent Cells}} & \multicolumn{4}{c}{\textbf{Adjacent Cells}} \\
            \cmidrule(lr){2-5} \cmidrule(lr){6-9}
            & \textbf{AJI} & \textbf{IoU} & \textbf{FN} & \textbf{FP} & \textbf{AJI} & \textbf{IoU} & \textbf{FN} & \textbf{FP} \\
            \midrule
            SSA \cite{ssa2020miccai} {\tiny MICCAI'20} & 0.201 & 0.547 & 0.254 & 0.199 & 0.176 & 0.546 & 0.267 & 0.188 \\
            $+$\textit{Step 1} (\cref{step1}) & 0.300 & 0.435 & 0.110 & 0.456 & 0.252 & 0.471 & 0.142 & 0.387 \\
            $+$\textit{Step 2} (\cref{step2}) & 0.341 & 0.615 & 0.235 & 0.149 & 0.211 & 0.529 & 0.325 & 0.146 \\
            \rowcolor{gray!20} \textbf{$+$\textit{Step 3} (\cref{step3})} & \textbf{0.510} & \textbf{0.663} & \textbf{0.126} & \textbf{0.211} & \textbf{0.405} & \textbf{0.701} & \textbf{0.152} & \textbf{0.147} \\
            \bottomrule
        \end{tabular}
    \end{threeparttable}
\end{table}

\begin{table}[t]
    \centering
    \caption{Effect of key components in COIN on the MoNuSeg train set \cite{kumar2017monuseg, kumar2020monuseg} (Baseline: SSA \cite{ssa2020miccai}, Extension of \cref{propagation_ablation}). }
    \label{rebutt_comp_ablation}
    \scriptsize
    \setlength{\tabcolsep}{3pt} 
    \renewcommand{\arraystretch}{1} 
    \begin{threeparttable}
    \begin{tabular}{l|lcccc|cccc}
        \toprule
        && \multicolumn{4}{c|}{\textbf{COIN Components}} & \multicolumn{4}{c}{\textbf{Metrics}} \\
        \cmidrule(lr){3-6} \cmidrule(lr){7-10}
        && \textbf{USS} & \textbf{OT} & \textbf{CRF} & \textbf{Watershed} & \textbf{AJI ($\uparrow$)} & \textbf{IoU ($\uparrow$)} & \textbf{FN ($\downarrow$)} & \textbf{FP ($\downarrow$)} \\
        \midrule
        %\smaller{(a)} & \smaller{Baseline} & \xmark & \xmark & \cmark & \cmark & - & - & - & - \\
        (a)& & \xmark & \xmark & \xmark & \xmark & 0.001 & 0.305 & \underline{0.536} & 0.159 \\
        \arrayrulecolor{lightgray} 
        % \midrule
        %\hdashline
        & & \cmark & \xmark & \xmark & \xmark & 0.001 & 0.439 & \textbf{0.020} & 0.552 \\
        \midrule
        %\hdashline
        (b)& & \cmark & \cmark & \xmark & \xmark & 0.001 & 0.539 & 0.163 & \textbf{0.303} \\
        & & \cmark & \xmark & \cmark & \xmark & 0.001 & 0.443 & 0.022 & \underline{0.549} \\
        \midrule
        %\hdashline
        (c) & & \cmark & \cmark & \cmark & \xmark & \underline{0.001} & 0.543 & 0.157 & 0.301 \\
        \rowcolor{gray!20} & \smaller{Ours/Step 1}& \cmark & \cmark & \cmark & \cmark & \textbf{0.380} & 0.543 & 0.157 & 0.301 \\
        \arrayrulecolor{black} 
        \bottomrule
    \end{tabular}  
    \end{threeparttable}
    \begin{tablenotes}
        \item +USS: FN($\downarrow$) \hspace{1cm} +OT and +CRF: FP($\downarrow$) \hspace{1cm} +Watershed: AJI($\uparrow$)
    \end{tablenotes}
\end{table}

Furthermore, in \cref{rebutt_comp_ablation}, we present a component-wise ablation study to individual modules of COIN and their contribution to the cell segmentation performance on the MoNuSeg \cite{kumar2017monuseg, kumar2020monuseg} train set. Compared to the baseline \cite{ssa2020miccai} without CRF and watershed (the first row), USS incorporation significantly reduced FN from 0.536 to 0.020 (the second row), accompanied by an increase in FP from 0.159 to 0.552 (\cref{rebutt_comp_ablation}(a)). While the application of CRF \cite{krahenbuhl2012crf} showed only a 0.4\%p increase in IoU (the fourth row), OT alone reduced FP by 1.8 times more than did CRF alone (\cref{rebutt_comp_ablation}(b)). Notably, applying OT before CRF reduces the FP from 0.552 to 0.303, which is almost identical to the reduction seen when applying OT alone (from 0.552 to 0.301), suggesting that OT is the key factor in adjusting FP, while CRF has minimal impact. Lastly, as shown in \cref{eq_step2}, watershed algorithm \cite{ssa2020miccai} separates adjacent binary masks into distinct instances, a standard post-processing step in all UCIS baselines \cite{ssa2020miccai, psm2023miccai}. Therefore, while IoU remained at 0.543, AJI increased from 0.001 to 0.380 (see \cref{rebutt_comp_ablation}(c)).

\section{Method Details}
\subsection{Details of Unsupervised Semantic Segmentation} \label{appendix_uss_details}
We are the first case to apply DINOv2 \cite{oquab2023dinov2} and MAE \cite{he2022masked} for analyzing pathological images (see \cref{overview}). As shown in \cref{USS+OT} and \cref{propagation_ablation}, the USS models \cite{oquab2023dinov2, he2022masked} group similar pixels (\emph{e.g.,} cells) from UCIS seeds \cite{ssa2020miccai}, {resulting in more than 26$\times$ reduction in FN} (see \cref{rebutt_comp_ablation}(b)). However, USS's pixel similarity-based grouping often fails to distinguish between cells and tissues of similar colors. As shown in \cref{OT_effect}, the USS output $S_{\theta}^{us}$ cannot differentiate cell activation from the background. We address this substantial increase in FP by incorporating optimal transport (OT) \cite{rachev1985monge} (see \cref{step1}, \cref{propagation_ablation}, and \cref{OT_effect}). %We emphasize that while we do utilize USS \cite{he2022masked}, our main contribution lies in how we integrate it with other pre-trained models.

\subsection{Class-level Average Pooling} \label{appendix_cap}
Class-level average pooling (CAP) \cite{jo2024dhr} is the modified version of the standard pooling technique (\emph{i.e.,} global average pooling) in which the average of the grouped embedding vectors outputs class-specific centroids. In \cref{step1}. the implementation of CAP to $M^{ucis}_\theta(I_k)$ yields class-specific USS centroids $V^{us}$. In our study, class denotes either cell or background. 

\subsection{Push Operation in Optimal Transport} \label{rebutt_push_operation}
{The push operation \(T\) involved in \cref{eq_S_OT} is the optimal-transport plan that redistributes the mass from the original pixel similarity distribution (\(S^{us}_{ij}\)) to the target distribution consisting of two distinct classes: foreground (cells) and background (tissue) \cite{jo2024dhr, li2023point2mask}. \(T_{ij}\) determines how much mass moves from pixel \(i\) to class \(j\) by minimizing \(\sum_{i,j} T_{ij}(1 - S^{us}_{ij}) - \lambda H(T))\).}

{The computed \(T\) then pushes the original similarity map \(S^{us}\) to the refined mask \(S^{OT}\) by \(S^{OT} = T \circ S^{us}\), sharpening pixel-wise foreground-background boundaries. We confirmed that this operation is robust to changes in $\lambda$ (\cref{appendix_ot_lambda}).}

\subsection{Watershed Algorithm} \label{appendix_watershed}
The watershed algorithm \cite{ssa2020miccai} is a classical image segmentation technique that is particularly effective for separating overlapping objects. Specifically, the image, treated like a topographic map, is turned into a grayscale that allows pixels to have distinctive values (0 to 255) with high intensity indicating peaks and low intensity denoting valleys. Imagine pouring water over this topographic map, where the valleys are flooded first and eventually merge as the water rises. Each valley contains different labels, and to prevent the labels from merging, the barriers are built at locations where water merges. This process continues until the peaks are all submerged underwater. Here, the barriers indicate the segmentation result. In previous work \cite{ssa2020miccai}, the instance is obtained in the post-processing step which uses the inverse of the distance transform and the local maxima as markers (\emph{i.e.,} labels) for the watershed algorithm (see Sec. 3.4 and Fig. 2 in \cite{ssa2020miccai}). Inspired by this, we utilize the watershed algorithm to obtain an initial instance mask $E^i_{{\theta_1}}(I_k)$ for $N$ instances before training the edge decoder in \cref{eq_step2}.
% \sj{In previous work \cite{ssa2020miccai}, the instance is obtained in the post-processing step which uses the inverse of the distance transform and the local maxima as markers (\emph{i.e.,} labels) for the watershed algorithm (see Sec. 3.4 and Fig. 2 in \cite{ssa2020miccai}). Inspired by this, we utilize the watershed algorithm to obtain initial instance mask $E^i_{{\theta_t}}(I_k)$ for $N$ instances before training the edge decoder (see \cref{step2} and \cref{eq_step2}).} 
%to delineate individual cells within the OT-refined similarity map $S^{OT}_{\theta}(I_k)$, which yields instance mask $E^i_{{\theta_t}}(I_k)$ for $N$ instances (see \cref{step2} and \cref{eq_step2}).

\subsection{Details of SAM Consistency} \label{appendix_sam}
{As shown in \cref{appendix_SAM_fig}, we hypothesize and confirm that SAM \cite{kirillov2023segment} faithfully reconstructs an instance's shape only when the input prompt (\emph{i.e.}, model-predicted mask) aligns closely with the ground truth, but when the prompt is noisy or incorrect, SAM often overgeneralizes and activates most of the surrounding pixels (\cref{appendix_SAM_fig}; SAM Failure Cases). Thus, our method does not rely solely on SAM because the application of SAM to out-of-distribution data (\emph{e.g.,} cell segmentation) itself introduces uncertainty. For example, when SAM randomly targets the background pixel, many pixels become overgeneralized as foreground, jeopardizing the segmentation performance (top right side of \cref{appendix_SAM_fig}). Therefore, a high IoU between the input prompt (model-predicted mask) and SAM's output reliably flags error-free instances, and these top-scoring masks achieve AJI values nearly identical to those using ground-truth labels (\cref{figure_correlation}). Specifically, COIN outputs low scores when either the UCIS baseline or SAM fails (right) and high scores when both succeed (left). Without our scoring approach, SAM would frequently assign multiple pixels in the background as cell instances, preventing the detection of individual cells. The high IoU scores corresponding to success cases for both predictions suggest that our instance-level confidence scoring method can automatically select highly confident instances for training without relying on ground truth annotations. Therefore, unlike the naïve application of SAM (the second row of \cref{sam_ablation}), we observe a substantial performance improvement when our scoring method is applied (the third row of \cref{sam_ablation}).}
% we observe - for the first time - a substantial performance improvement when our scoring method is implemented (third row of \cref{sam_ablation}).}
%Our method does not simply rely on SAM because the application of SAM \cite{kirillov2023segment} itself to out-of-distribution data (\emph{e.g.,} cell segmentation) causes uncertainty, especially when SAM randomly targets the background pixel. In this case, many pixels are designated as foreground, jeopardizing the segmentation performance (top right side of \cref{appendix_SAM_fig}). As depicted in \cref{appendix_SAM_fig}, we present the IoU scores corresponding to the success and failure cases of SAM and the UCIS baseline \cite{ssa2020miccai}. Specifically, COIN outputs low scores when either the UCIS baseline or SAM fails (right) and high scores when both succeed (left). Without our scoring approach, SAM would frequently assign multiple pixels in the background as cell instances, preventing the detection of individual cells. The high IoU scores corresponding to success cases for both predictions suggest that our instance-level confidence scoring method can automatically select highly confident instances for training without relying on ground truth annotations. Therefore, unlike the naïve application of SAM (the second row of \cref{sam_ablation}), we observe a substantial performance improvement when our scoring method is applied (the third row of \cref{sam_ablation}). % we observe - for the first time - a substantial performance improvement when our scoring method is implemented (third row of \cref{sam_ablation}).

\subsection{Canny Algorithm} \label{appendix_canny}
In contrast to standard edge detection applications that process RGB images, we simply extract edges from binary masks (see \cref{eq_edge}). Therefore, we utilize the traditional and well-known Canny algorithm \cite{canny1986computational}. Processing a $1000 \times 1000$ binary mask with this algorithm requires approximately seven milliseconds.
% \sj{In contrast to standard edge detection applications that process RGB images, we simply extract edges from binary masks (see \cref{eq_edge}). Therefore, we utilize the traditional and well-known Canny edge detection algorithm \cite{canny1986computational}. Processing a $1000 \times 1000$ binary mask with this algorithm requires approximately seven milliseconds.}
%Canny Edge Detection \cite{canny1986computational} algorithm is a widely used technique for detecting image edges. First, the algorithm applies a $5\times5$ Gaussian filter, resulting in a slightly blurred image with reduced noise. Second, to identify regions with high spatial derivatives that correspond to edges, the gradient is calculated in both x and y directions using edge detection operators (\emph{e.g.,} Sobel filter). This process highlights regions with notable intensity changes and determines the orientation of edges. Third, to specify the edges, the gradient magnitude of two neighboring pixels along the gradient direction is compared. If the pixel's magnitude does not exceed that of both neighbors, it is set to zero. This process produces edges that are more thin and precise. Fourth, the edges are categorized as strong, weak, or non-relevant based on gradient magnitudes by thresholding to distinguish important edges. Lastly, the weak edges connected to strong edges are retained, ensuring continuity of edges. Those that are not connected to strong edges are discarded. Our method utilizes Canny algorithm to create pseudo edge mask $\hat{M}_{edge}^i(t)$, as indicated in \cref{eq_edge}.

\subsection{Details of Pseudo Masks and Edge Decoder} \label{appendix_edge}
In \emph{Step 3} (\cref{step3}), two pseudo masks are generated based on the accepted indices $\mathcal{A}_{\delta}$ from \cref{eq_A}. As depicted in \cref{appendix_pseudomask_fig}, pseudo binary mask $\hat{M}_{bin}^i(t)$ from \cref{eq_bin} refers to the pixels designated as foreground (\emph{i.e.,} cell), which corresponds to high-scoring instances within the scored instances. Therefore, the low-scoring instances are omitted and not used for training. The pseudo edge mask $\hat{M}_{edge}^i(t)$ from \cref{eq_edge} denotes the cell boundaries. The pseudo binary and edge masks in \cref{appendix_pseudomask_fig} are the decomposed representation of the pseudo mask at $t=1$ from \cref{Recursive}.

Unlike existing UCIS models \cite{ssa2020miccai, psm2023miccai}, our framework incorporates an edge decoder to train on pseudo edge masks. Inspired by recent studies \cite{li2023point2mask, peng2020deepsnake, liang2021polytransform}, the edge decoder learns the boundaries between neighboring instances to address the challenge of distinguishing adjacent cells. Specifically, DeepSnake \cite{peng2020deepsnake} trains on the loss from iterative contour deformation (refer to Eq. (4) at \cite{peng2020deepsnake}), which iteratively deforms the initial contour to approach the actual object boundary, and Point2Mask \cite{li2023point2mask} learns high-level boundary map by utilizing the mask affinity equivalence among the eight neighbor pixels (refer to Eq. (7) at \cite{li2023point2mask}). {PolyTransform \cite{liang2021polytransform} trains on the losses from the feature extraction network and deforming network for learning strong object boundaries and predicting the offset for each vertex, respectively (refer to Sec. 3.4 from \cite{liang2021polytransform}).} Thus, including an edge decoder allows our approach to learn discriminative instance features during training, leading to significant improvements in segmentation accuracy (see \cref{appendix_edgedecoder_table}).
%PolyTransform \cite{liang2021polytransform} trains on the losses from the feature extraction network, for learning strong object boundaries, and the deforming network, for predicting the offset for each vertex (refer to Sec. 3.4 from \cite{liang2021polytransform}). Thus, including an edge decoder allows our approach to learn discriminative edge features during training, leading to significant improvements in segmentation accuracy (\cref{appendix_edgedecoder}).

\begin{figure}[t]
  \centering
  \includegraphics[width=1.0\linewidth]{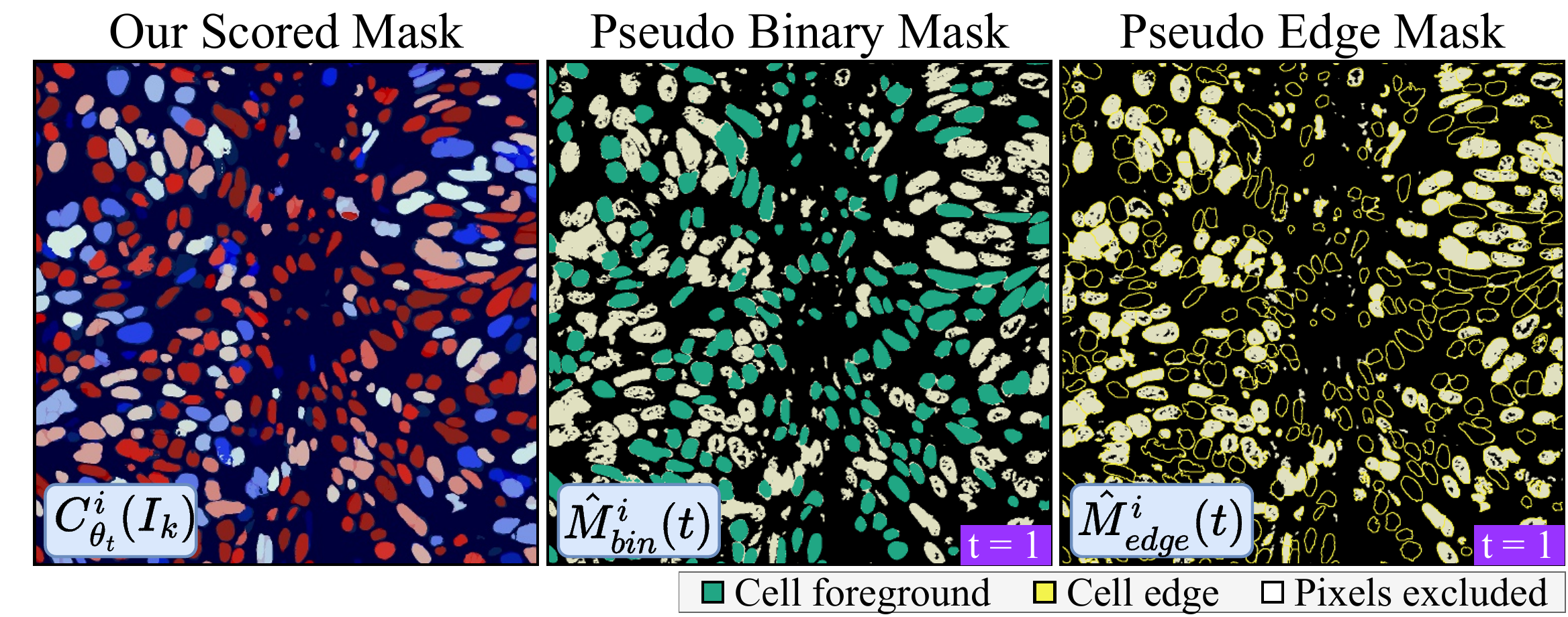}
  \caption{\textbf{Illustration of binary and edge pseudo masks.} Green represents the cell foreground, and yellow lines denote the cell edges. White indicate pixels excluded from training. Note that only high-scoring instances (red) are used to generate pseudo masks.}
  \label{appendix_pseudomask_fig}
\end{figure} 

\subsection{Datasets} \label{appendix_dataset}
Main experiments (\cref{main_table}) are conducted on MoNuSeg \cite{kumar2020monuseg, kumar2017monuseg} and TNBC \cite{naylor2019tnbc} datasets.  MoNuSeg contains multi-organ nuclei segmentation images that are H\&E-stained and captured at 40x magnification. Specifically, it includes a total of 21,623 annotated nuclear boundaries. TNBC (Triple Negative Breast Cancer) dataset is generated at the Curie Institute and consists of 50 images with 4,022 annotated cells. BRCA \cite{dataset2021brca} contains breast cancer H\&E-stained images. CPM-17 \cite{vu2018cpm} and PanNuke \cite{gamper2020pannuke} are derived from multiple types of tissues, consisting of 205,343 and 7,750 annotated nuclei, respectively. CryoNuSeg \cite{dataset2021cryonuseg} contains fully annotated H\&E-stained nuclei instance segmentation images derived from frozen tissue samples (FS) of 10 human organs.

\section{Additional Quantitative Results}
\subsection{OT Hyperparameters}
The experiment on OT parameters, as shown in \cref{appendix_ot_lambda}, demonstrates that the IoU values remain stable across varying $\lambda$ values. Specifically, when $\lambda$ is increased from 0.01 to 0.4, the IoU fluctuates only slightly, with the highest IoU observed at $\lambda = 0.1$ (0.543) and the lowest at $\lambda = 0.01$ (0.532). The difference between the maximum and minimum IoU is just 0.011, indicating that the model's performance is not significantly influenced by the parameter value.

\begin{figure}[t]
  \centering
  \includegraphics[width=0.8\linewidth]{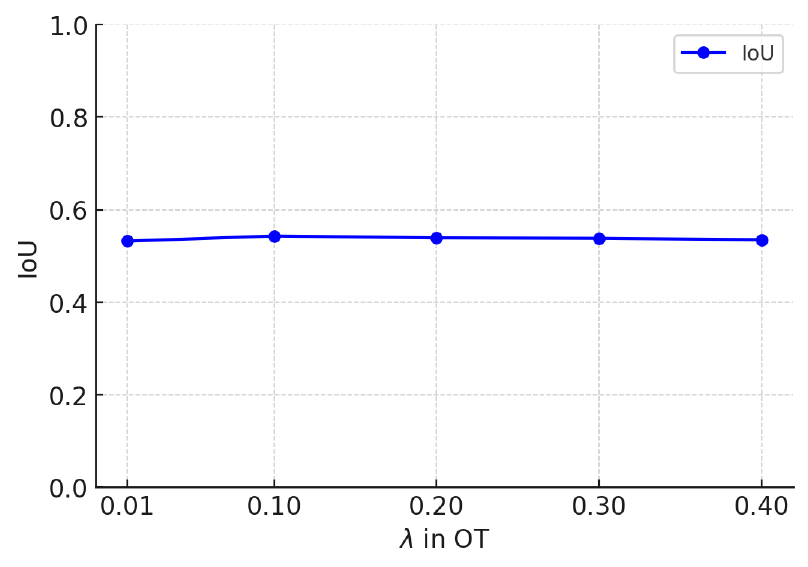}
  \vspace{-0.35cm}
  \caption{\textbf{Performance analysis with varying OT parameters.}}
  \label{appendix_ot_lambda}
\end{figure} 

\begin{figure}[t]
  \centering
  \includegraphics[width=0.95\linewidth]{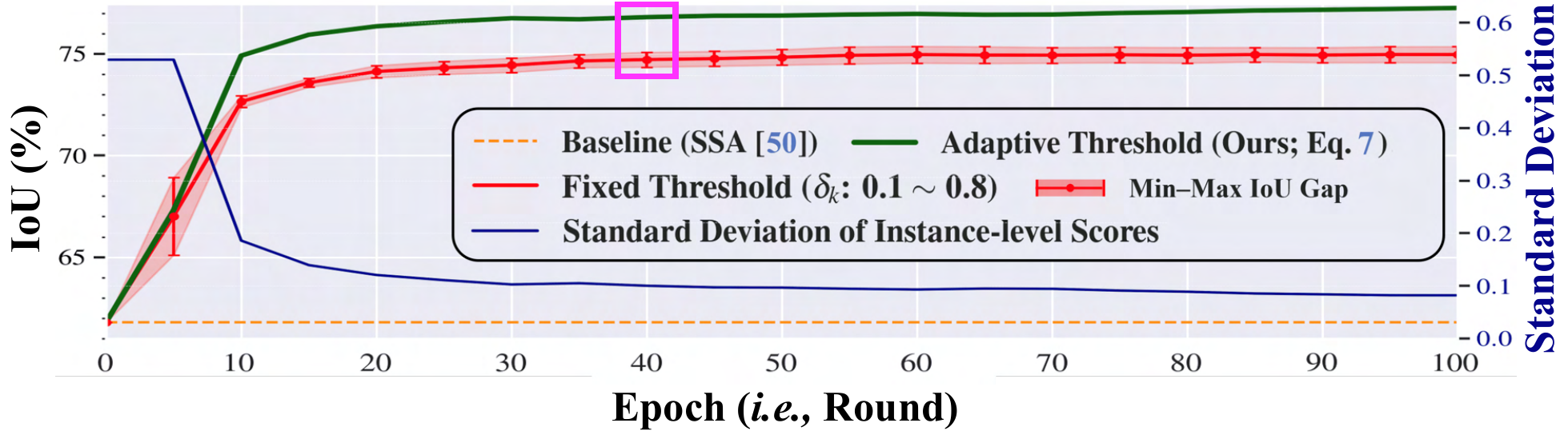}
  \vspace{-0.35cm}
  \caption{\textbf{Performance comparison between fixed and adaptive threshold.}}
  \label{appendix_rebutt_graph}
\end{figure} 

\subsection{Effect of Adaptive Thresholding}
{We train for 100 epochs as in all experiments (\cref{rebutt_comp_complexity}) and observe that on the MoNuSeg \cite{kumar2017monuseg, kumar2020monuseg} test set, IoU steadily improves and plateaus around 40\% of training (2.4 hours; \cref{appendix_rebutt_graph}, pink box), regardless of threshold type. Notably, our non-parametric, adaptive threshold (green) consistently outperforms fixed parametric variants by about 3\%p in IoU.}

{To further understand why $\delta_k$ adapts so effectively, we plot the standard deviation of consistency scores across training epochs (\cref{appendix_rebutt_graph}, blue line). Since $\delta_k$ is characterized by dividing its standard deviation by the mean, and a high standard deviation implies a large variation in predicted instance scores (\emph{i.e.,} greater uncertainty), we focus on how this uncertainty evolves during training. As shown in \cref{appendix_rebutt_graph}, the standard deviation is high early in the training process, indicating that model predictions vary significantly due to noisy or uncertain instances. However, as training progresses, the standard deviation steadily decreases, reflecting a stabilization of model predictions. This aligns with the notion of performance saturation (pink box). The adaptive nature of $\delta_k$ enables it to respond to these changes by filtering out early noise and adjusting as confidence solidifies. In contrast, the fixed thresholds fail to remove noisy instances effectively in the early stages of self-distillation, resulting in suboptimal performance compared to our adaptive threshold.}
%\sj{Additionally, to examine why $\delta_k$ adapts so effectively, we plot the standard deviation of the consistency scores across epochs (\cref{appendix_rebutt_graph}, blue line). Initial high deviations indicate inconsistent model-SAM outputs due to uncertainty. With each iteration, the deviation decreases progressively, reflecting the stabilization of the model predictions. Thus, $\delta_k$ self-adjusts, filtering out early noise and adapting as the model's confidence solidifies.}

\subsection{Model-agnostic Improvements with Various USS Backbones} \label{appendix_uss}
We extensively evaluate our method by experimenting on different USS backbones \cite{oquab2023dinov2, he2022masked, caron2021emerging, darcet2023vision} in all metrics on TNBC \cite{naylor2019tnbc} test set. In \cref{appendix_uss_table}, we compare the performance of Masked Autoencoder (MAE) \cite{he2022masked}, which is leveraged for all other experiments, against DINOv1 \cite{caron2021emerging}, DINOv2 \cite{oquab2023dinov2}, and DINOv2-reg \cite{darcet2023vision}, which demonstrates that MAE outperforms all USS backbones across all metrics. Specifically, for instance segmentation performance, MAE outperforms DINOv1 by at least $+$2\%p, DINOv2 by at least $+$1\%p, and DINOv2-reg by at least $+$0.5\%p. MAE also surpasses other USS backbones regarding semantic segmentation, with at least $+$3.3\%p for DINOv1, $+$2.6\%p for DINOv2, and $+$1.7\%p for DINOv2-reg. Therefore, we select MAE as the USS backbone for all experiments. 

\begin{table}[t]
    \setlength{\tabcolsep}{8pt}
    \renewcommand{\arraystretch}{1.2} 
    \caption{Comparison of four USS backbones \cite{caron2021emerging, oquab2023dinov2, darcet2023vision, he2022masked} on the TNBC \cite{naylor2019tnbc} test set.}
    \vspace{-0.35cm}
    \label{appendix_uss_table}
    \scriptsize
    \centering
    \setlength{\tabcolsep}{8pt}
    \renewcommand{\arraystretch}{1.0}
    \begin{threeparttable}
        \begin{tabular}{l|c c c c} 
            \toprule
            {\textbf{Backbone}} & \multicolumn{2}{c}{\textbf{Instance Segmentation}} & \multicolumn{2}{c}{\textbf{Semantic Segmentation}} \\
            \cmidrule(lr){2-3} \cmidrule(lr){4-5}
             & \textbf{AJI} & \textbf{PQ} & \textbf{IoU} & \textbf{Dice} \\
            \midrule
            DINOv1 \cite{caron2021emerging} & 0.534 & 0.519 & 0.764 & 0.721\\
            DINOv2 \cite{oquab2023dinov2} & 0.558 & 0.528 & 0.771 & 0.733\\
            DINOv2-reg \cite{darcet2023vision}  & 0.563 & 0.533 & 0.780 & 0.754\\
            \rowcolor{gray!20} MAE \cite{he2022masked} & \textbf{0.568} & \textbf{0.540} & \textbf{0.797} & \textbf{0.774}\\
            \bottomrule
        \end{tabular}
    \end{threeparttable}
\end{table}

\begin{table}[t]
    \caption{Performance evaluation of COIN on adjacent and non-adjacent cells on the MoNuSeg \cite{kumar2017monuseg, kumar2020monuseg} test set.}
    \vspace{-0.35cm}
    \label{appendix_adjacent_table}
    \scriptsize
    \setlength{\tabcolsep}{9pt}
    \centering
    \renewcommand{\arraystretch}{1.0} 
    \begin{threeparttable}
    \begin{tabular}{lcccc}
        \toprule
        \textbf{Method} & \multicolumn{2}{c}{\textbf{Non-adjacent Cells}} & \multicolumn{2}{c}{\textbf{Adjacent Cells}} \\
        \cmidrule(lr){2-3} \cmidrule(lr){4-5}
         & \textbf{AJI} ($\uparrow$)& \textbf{IoU} ($\uparrow$)& \textbf{AJI} ($\uparrow$)& \textbf{IoU} ($\uparrow$)\\
        \midrule
        SSA \cite{ssa2020miccai} {\tiny MICCAI'20} & 0.288 & 0.583 & 0.235 & 0.632 \\
        \rowcolor{gray!20} \textbf{SSA + COIN (Ours)} & \textbf{0.602} & \textbf{0.729} & \textbf{0.528} & \textbf{0.750} \\
        \rowcolor{gray!20} $\Delta_{ssa}$ & \textcolor{blue}{+0.314} & \textcolor{blue}{+0.146} & \textcolor{blue}{+0.293} & \textcolor{blue}{+0.118} \\
        \midrule
        PSM \cite{psm2023miccai} {\tiny MICCAI'23} & 0.498 & 0.695 & 0.408 & 0.660 \\
        \rowcolor{gray!20} \textbf{PSM + COIN (Ours)} & \textbf{0.601} & \textbf{0.725} & \textbf{0.527} & \textbf{0.748} \\
        \rowcolor{gray!20} $\Delta_{psm}$ & \textcolor{blue}{+0.103} & \textcolor{blue}{+0.030} & \textcolor{blue}{+0.119} & \textcolor{blue}{+0.088} \\
        \bottomrule
    \end{tabular}
    \begin{tablenotes}
        \item $\Delta_{ssa}$: Performance gap between SSA \cite{ssa2020miccai} and our proposed method.
        \item $\Delta_{psm}$: Performance gap between PSM \cite{psm2023miccai} and our proposed method.    
    \end{tablenotes}
    \end{threeparttable}
\end{table}

\subsection{Performance on Adjacent Cells} \label{appendix_adjacent}
To validate our method's performance in distinguishing adjacent cells, we specifically categorize non-adjacent cells and adjacent cells in the ground truth image of MoNuSeg \cite{kumar2017monuseg, kumar2020monuseg}, as depicted in \cref{appendix_adjacentcell_fig}. Following the dilation of ground truth cell edges and connected component labeling (CCL) \cite{rosenfeld1966sequential}, we identify cells that are connected to two or more cells as adjacent cells. \cref{appendix_adjacent_table} demonstrates that our approach consistently enhances the performances of existing UCIS models across both adjacent and non-adjacent cell types. For non-adjacent cells, our model achieves $+$31.4\%p in AJI and $+$14.6\%p in IoU with SSA and $+$10.3\%p in AJI and $+$3\%p in IoU with PSM. Notably, a similar pattern of performance improvement occurs with adjacent cells, with $+$29.3\%p in AJI and $+$11.8\%p in IoU with SSA and $+$11.9\%p in AJI and $+$8.8\%p in IoU with PSM. These results highlight COIN's ability to accurately separate instances, validating performance improvements in cell instance segmentation.

\begin{figure}[t]
  \centering
  \includegraphics[width=0.95\linewidth]{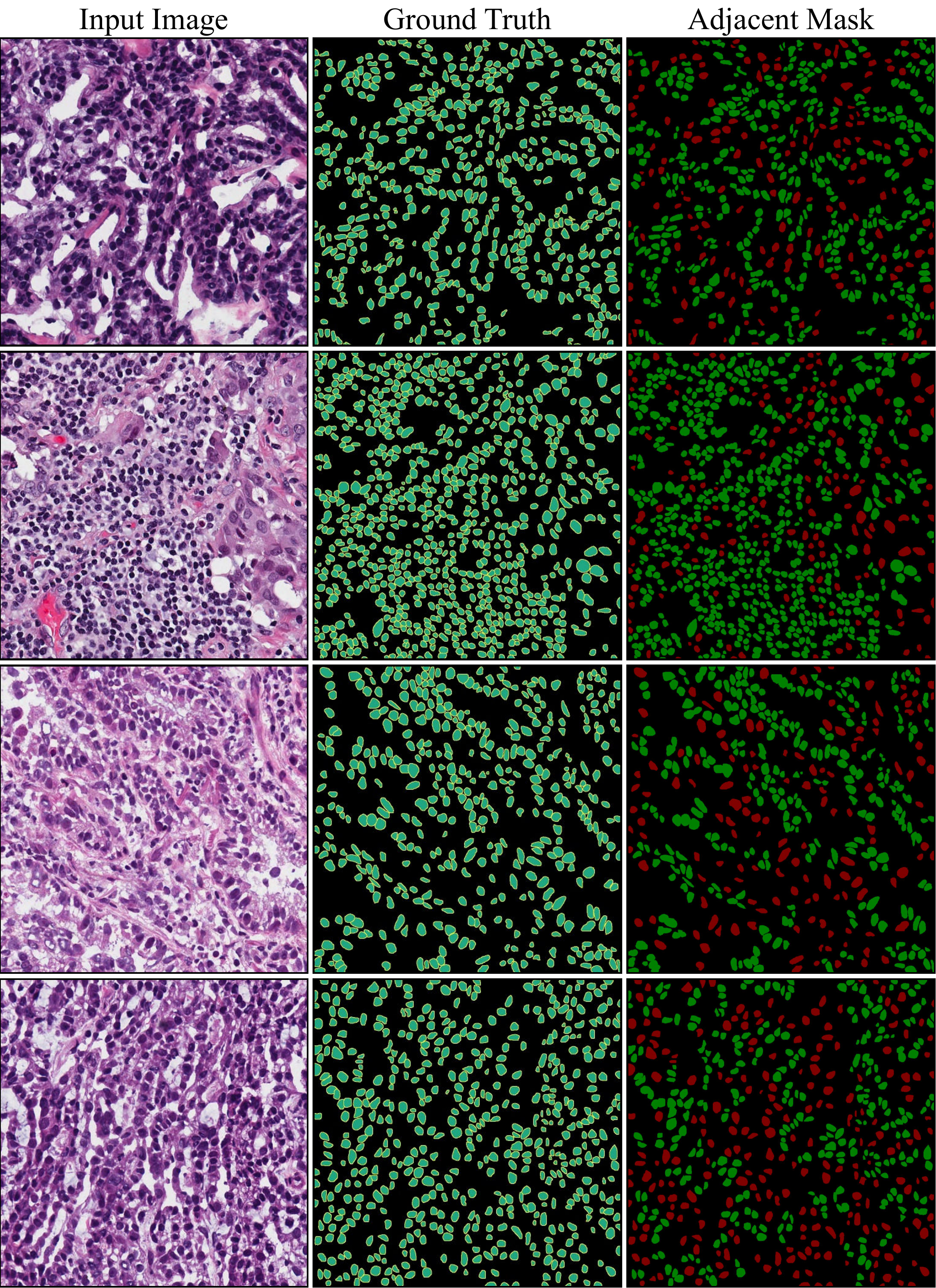}
  \vspace{-0.35cm}
  \caption{\textbf{Illustration of adjacent and non-adjacent cells.} Green represents the adjacent cells that are connected to at least two other cells, and red indicates non-adjacent cells that are not connected to any other cells.}
  \label{appendix_adjacentcell_fig}
\end{figure} 

\begin{table}[t]
    \centering
    \caption{Effect of the edge decoder on the MoNuSeg \cite{kumar2017monuseg, kumar2020monuseg} test set.}
    \label{appendix_edgedecoder_table}
    \scriptsize
    \setlength{\tabcolsep}{11pt} 
    \renewcommand{\arraystretch}{1.0} 
    \begin{threeparttable}
    \begin{tabular}{cccccc}
        \toprule
        {\textbf{Edge Decoder}} & \multicolumn{2}{c}{\textbf{Non-adjacent Cells}} & \multicolumn{2}{c}{\textbf{Adjacent Cells}} \\
        \cmidrule(lr){2-3} \cmidrule(lr){4-5}
        & \textbf{AJI} ($\uparrow$)& \textbf{IoU} ($\uparrow$)& \textbf{AJI} ($\uparrow$)& \textbf{IoU} ($\uparrow$)\\
        \midrule
        \xmark & 0.594 & 0.712 & 0.493 & 0.738 \\
        \rowcolor{gray!20} \cmark & \textbf{0.602} & \textbf{0.729} & \textbf{0.528} & \textbf{0.750} \\
        \rowcolor{gray!20} $\Delta_{edge}$ & \textcolor{blue}{+0.008} & \textcolor{blue}{+0.017} & \textcolor{blue}{+0.035} & \textcolor{blue}{+0.012} \\
        \bottomrule
    \end{tabular}
    \begin{tablenotes}
        \item $\Delta_{edge}$: Performance enhancement made by training the edge decoder.
    \end{tablenotes}
    \end{threeparttable}
\end{table}

\subsection{Effect of Edge Decoder} \label{appendix_edgedecoder}
{In \cref{appendix_edgedecoder_table}, we assess the impact of incorporating an edge decoder on the MoNuSeg \cite{kumar2017monuseg, kumar2020monuseg} test set. The application of our edge decoder enhances segmentation performance for both adjacent and non-adjacent cells. Notably, for adjacent cells, the performance improves by $+$1.7\%p in AJI and $+$3.5\%p in IoU when the edge decoder is present, demonstrating its capability to effectively distinguish cell boundaries and enhance overall segmentation results.}

\section{Additional Qualitative Results}
\subsection{Comparison of UIS Methods and Ours} \label{appendix_uis}
We compare the qualitative performance of our method against UIS baselines \cite{wang2023cutler, li2024promerge} in \cref{appendix_uis_fig}. As demonstrated in \cref{main_table}, our proposed method substantially outperforms all UIS models \cite{caron2021emerging, oquab2023dinov2, he2022masked} that have low AJI scores. 

\subsection{Examples of SAM-based Instance-level Confidence Scoring} \label{appendix_scoring}
Previous methods \cite{xu2023sppnet, nam2024instasam} that leverage SAM \cite{kirillov2023segment} rely on outputs generated by SAM from manual annotations (\emph{e.g.,} points) to create pseudo labels. Their dependency on such annotations indicates that the annotation burden is persistent. In contrast, our work utilizes SAM for confidence measurement and confident instance selection without requiring SAM-based image-related manual annotations (see \cref{step2}). The proposed scoring process is completely unsupervised and automatic, and it is the first-ever case to leverage SAM for confidence score-related tasks. Refer to \cref{appendix_Scoring_fig} for example visualizations of SAM-based scoring. Notably, our scoring approach separates adjacent cells effectively even when the pseudo mask doesn't distinguish individual cells (the fourth row). 

\subsection{Limitations of Recursive Self-distillation} \label{appendix_transparent_failure}
{When we tracked IoU across each self-distillation iteration in \cref{appendix_rebutt_graph}, we noticed that a small number of noisy pseudo-labels persist in rare cases when the initial USS propagation fails (\emph{e.g.,} transparent cells). This particular case is a persistent challenge faced by prior UCIS approaches \cite{ssa2020miccai, psm2023miccai} including ours, but nonetheless, these cases are rare and represent only a small fraction of our datasets, exerting limited influence on the overall accuracy. We demonstrate example failure cases in \cref{appendix_distill_fail}.}

\subsection{Additional State-of-the-art Qualitative Results} \label{appendix_mainresults}
In Figs. \ref{appendix_monuseg} and \ref{appendix_tnbc}, we provide additional qualitative comparisons between our COIN method, two image-related annotation-driven models \cite{feng2024textdiff, xu2023sppnet}, and one image-related annotation-free model \cite{ssa2020miccai}. As confirmed by the improvements across all metrics in \cref{main_table}, these visual examples further highlight that the output of our method is not only comparable but often surpasses the performance of supervised models that depend on image-related annotations. It is noteworthy given that COIN achieves high-quality segmentation without depending on such labor-intensive and time-consuming annotations.

\subsection{Model-agnostic Improvements with Various UCIS Models} \label{appendix_modelagnostic}
\cref{appendix_modelagnostic_figure} illustrates the model-agnostic performance improvement by COIN on two different UCIS baselines \cite{ssa2020miccai, psm2023miccai}. Our framework notably improves the segmentation performance for both SSA \cite{ssa2020miccai} and PSM \cite{psm2023miccai}, demonstrating its model-agnostic nature. Specifically, COIN significantly improves SSA's missed and chunky predictions and PSM's incomplete edges, demonstrating the flexibility of our method. 

\subsection{Consistent Improvements on Multiple Datasets} \label{appendix_datasets}
In Figs. \ref{appendix_ModelAgnostic}, \ref{appendix_brca}, \ref{appendix_cpm}, \ref{appendix_cryonuseg}, and \ref{appendix_pannuke}, we further validate the scalability of our method by comparing qualitative improvements against the UCIS baseline (\emph{e.g.,} SSA \cite{ssa2020miccai}) on multiple datasets, including BRCA \cite{dataset2021brca}, CPM-17 \cite{vu2018cpm}, CryoNuSeg \cite{dataset2021cryonuseg}, and PanNuke \cite{gamper2020pannuke}. As demonstrated in \cref{table_ModelAgnostic}, our model combined with SSA substantially improves semantic and instance segmentation performances throughout multiple datasets. 

%%%%
% SAM-based scoring
\begin{figure*}[p]
  \centering
  \includegraphics[width=0.9\textwidth]{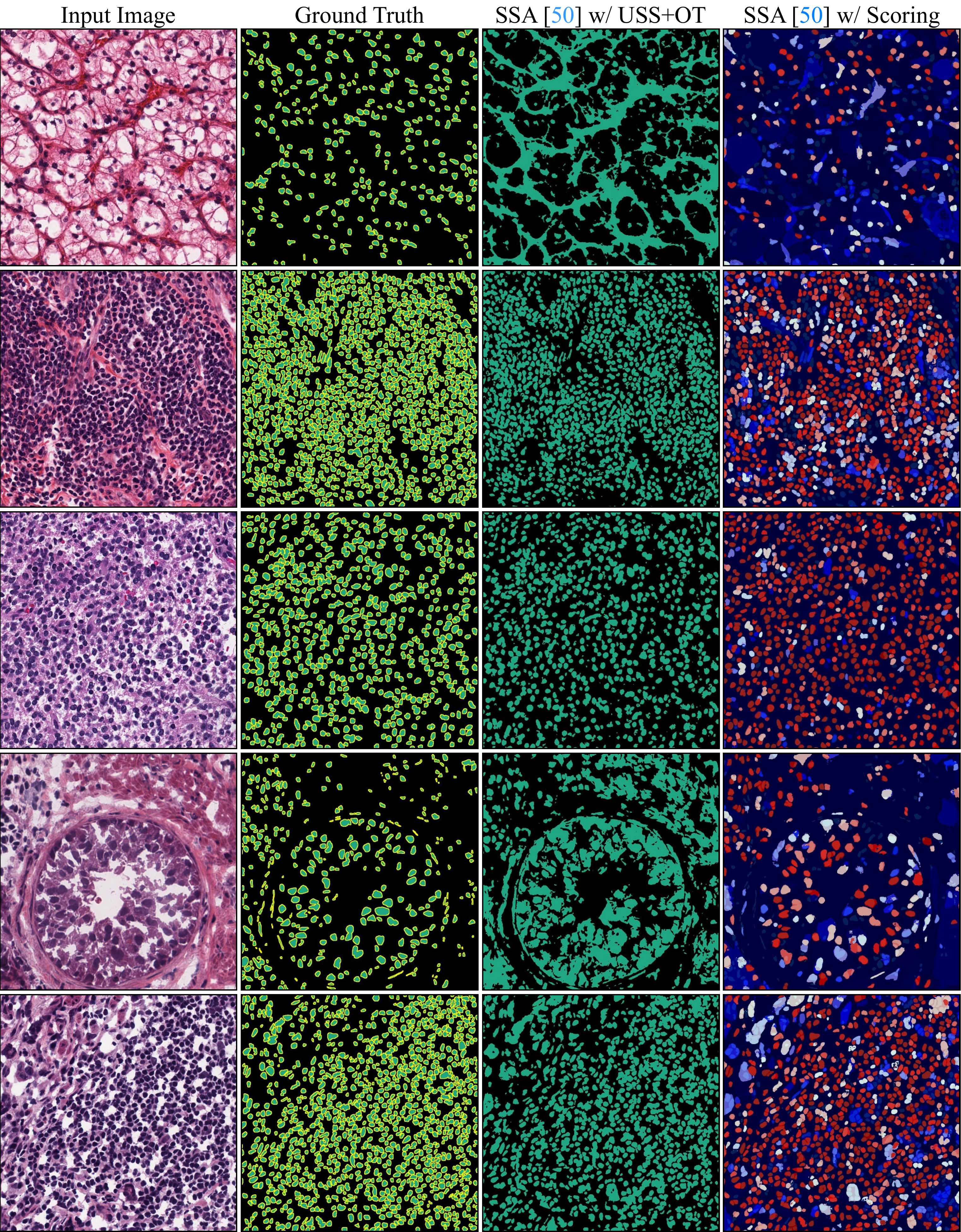}
  \caption{Qualitative examples of our instance-level confidence scoring based on SAM \cite{kirillov2023segment}.}
  \label{appendix_Scoring_fig}
\end{figure*}

% UIS
\begin{figure*}[p]
  \centering
  \includegraphics[width=0.8\linewidth]{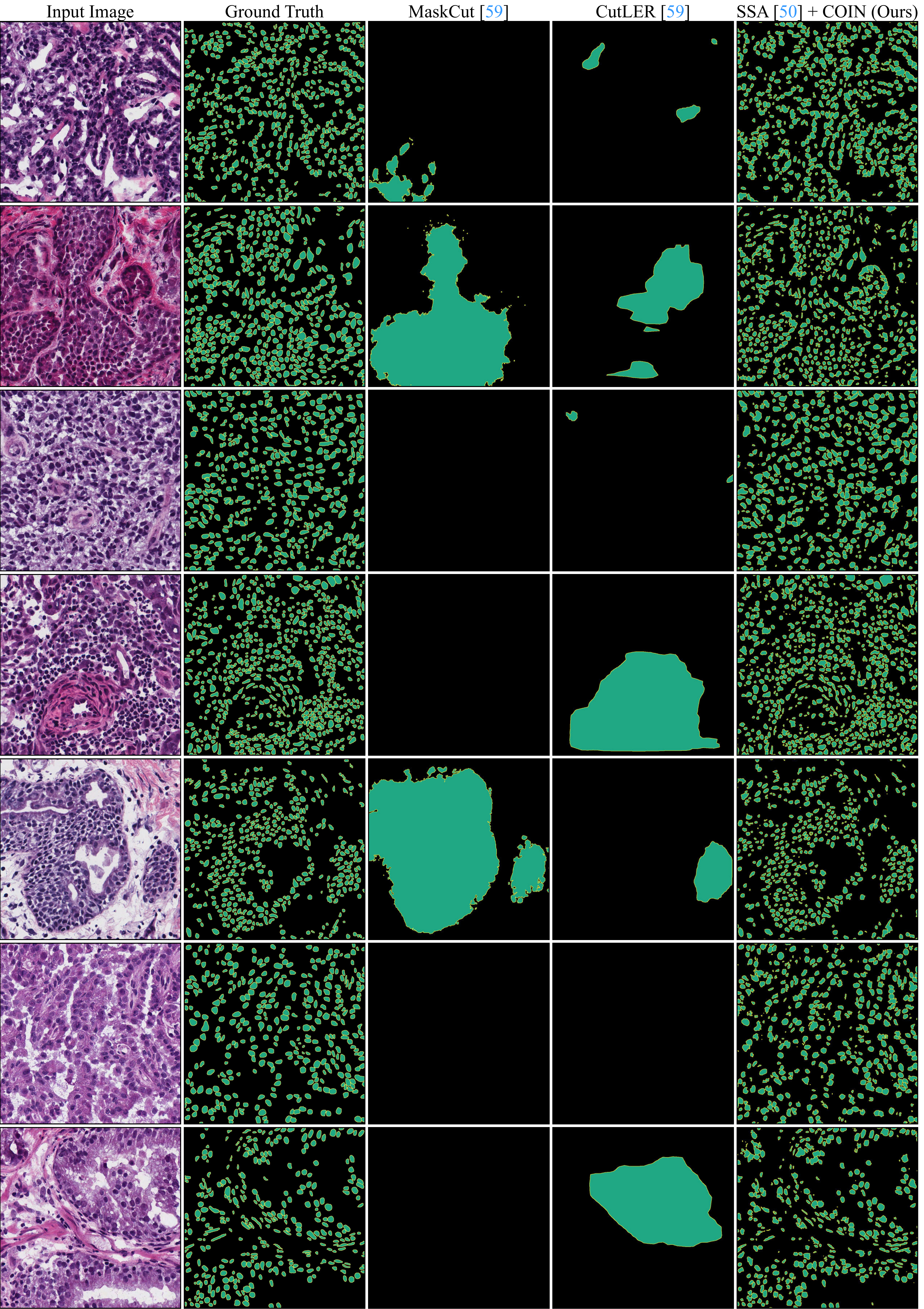}
  \caption{Qualitative comparison of UIS \cite{wang2023cutler} and COIN combined with SSA \cite{ssa2020miccai} on the MoNuSeg \cite{kumar2017monuseg, kumar2020monuseg} test set.} 
  \label{appendix_uis_fig}
\end{figure*} 

% SAM and model predictions
\begin{figure*}[p]
  \centering
  \includegraphics[width=0.95\textwidth]{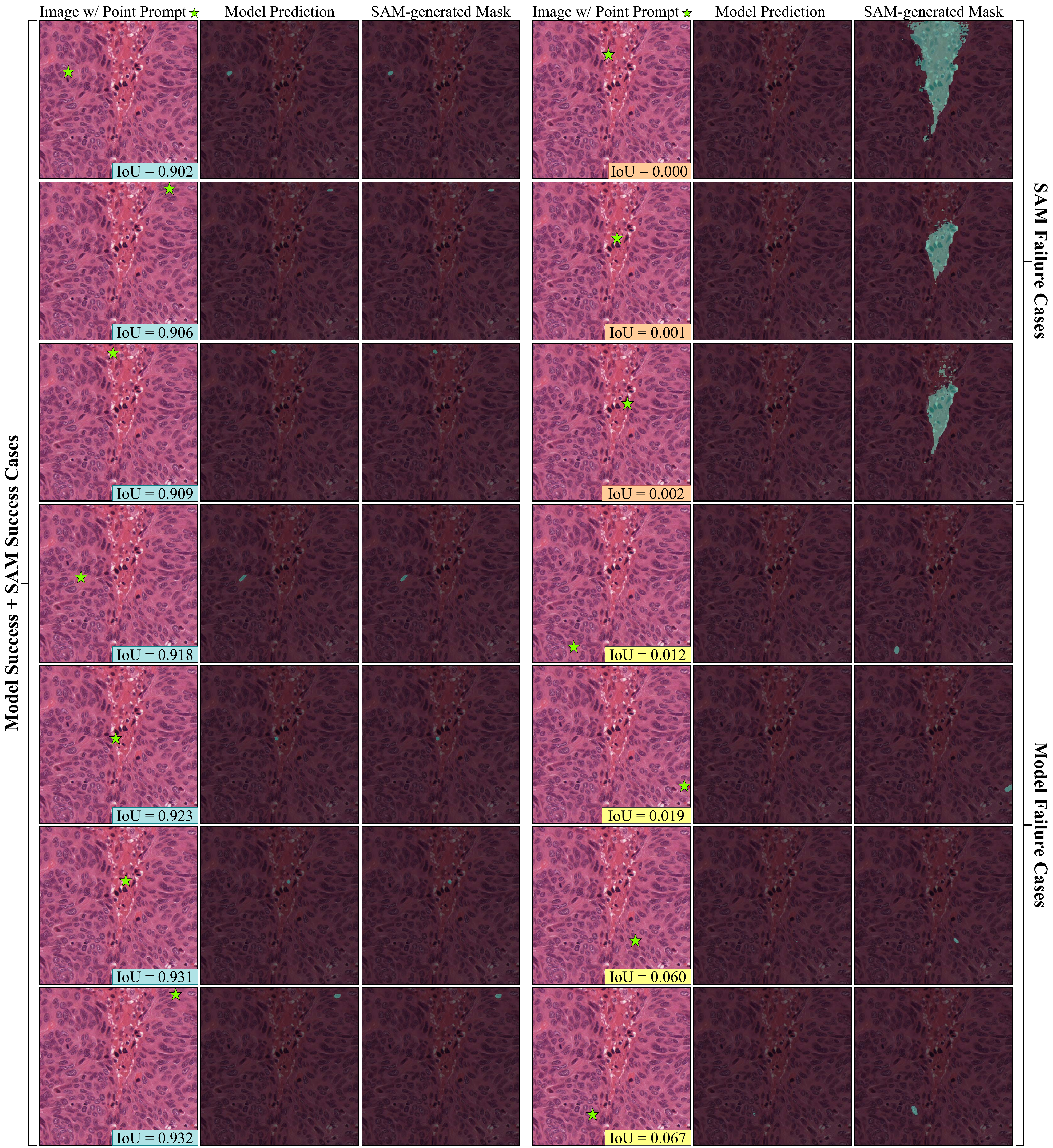}
  \caption{Visualization of success and failure cases for our propagated masks and their corresponding SAM-refined masks \cite{kirillov2023segment}.}
  \label{appendix_SAM_fig}
\end{figure*}

% Distillation failure case
\begin{figure*}[p]
  \centering
  \includegraphics[width=0.95\textwidth]{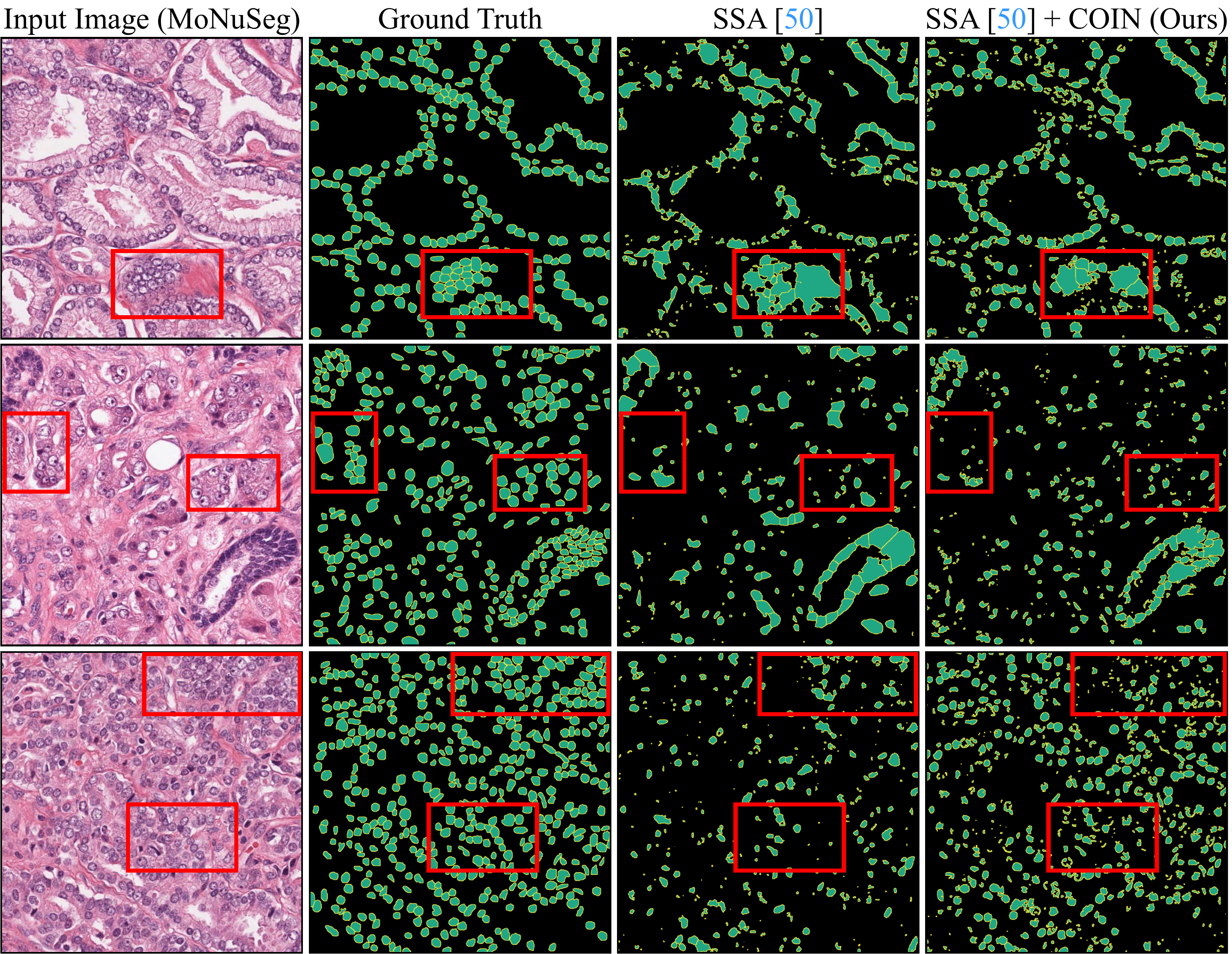}
  \caption{Visualization of failure cases for recursive self-distillation on the MoNuSeg \cite{kumar2017monuseg, kumar2020monuseg} train set.}
  \label{appendix_distill_fail}
\end{figure*}

% Main results qualitative results
\begin{figure*}[p]
  \centering
  \includegraphics[width=0.95\textwidth]{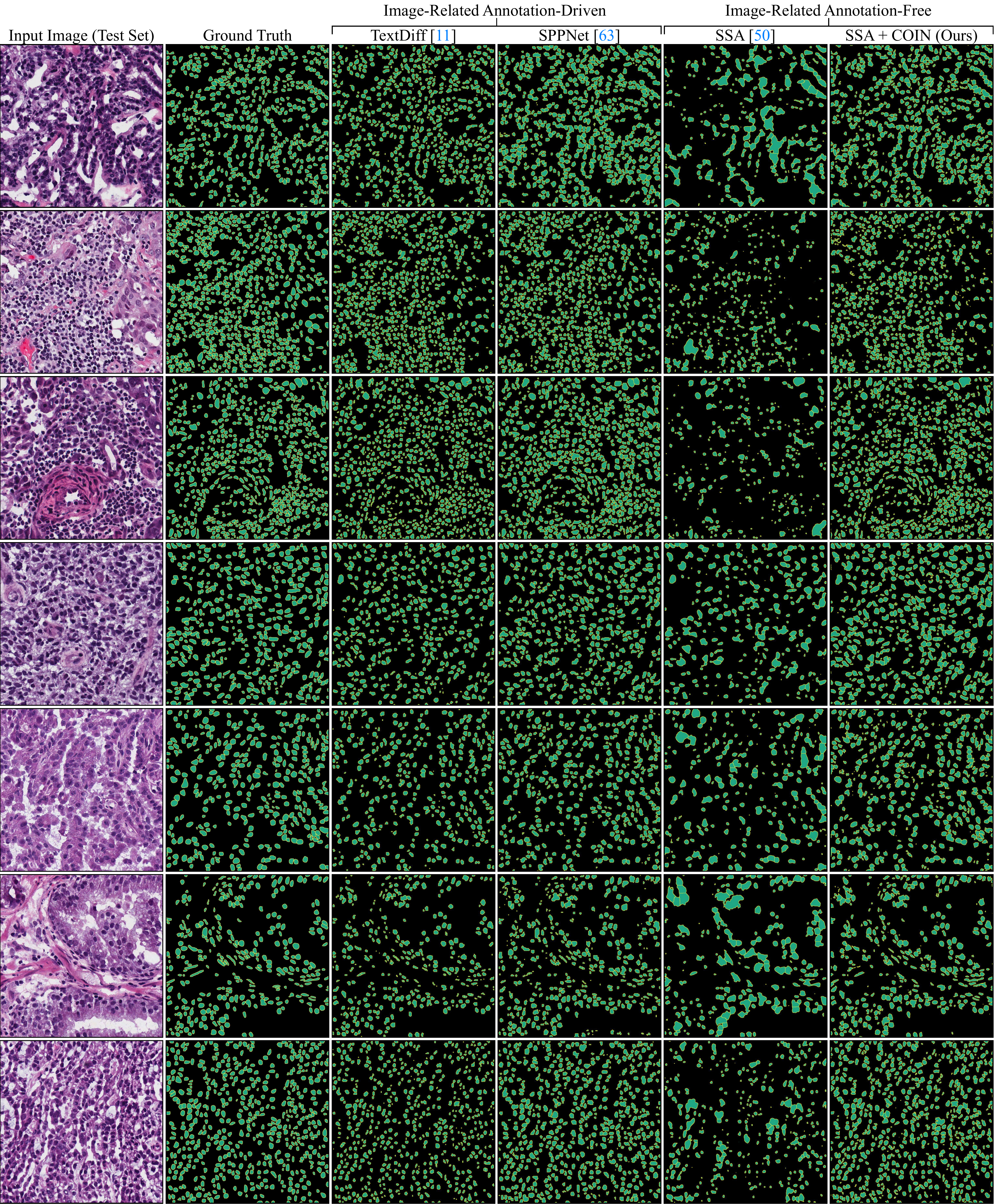}
  \caption{Qualitative comparison of annotation-driven and -free methods \cite{ssa2020miccai, feng2024textdiff, xu2023sppnet} on the MoNuSeg \cite{kumar2017monuseg, kumar2020monuseg} test set.}
  \label{appendix_monuseg}
\end{figure*}

\begin{figure*}[p]
  \centering
  \includegraphics[width=0.95\textwidth]{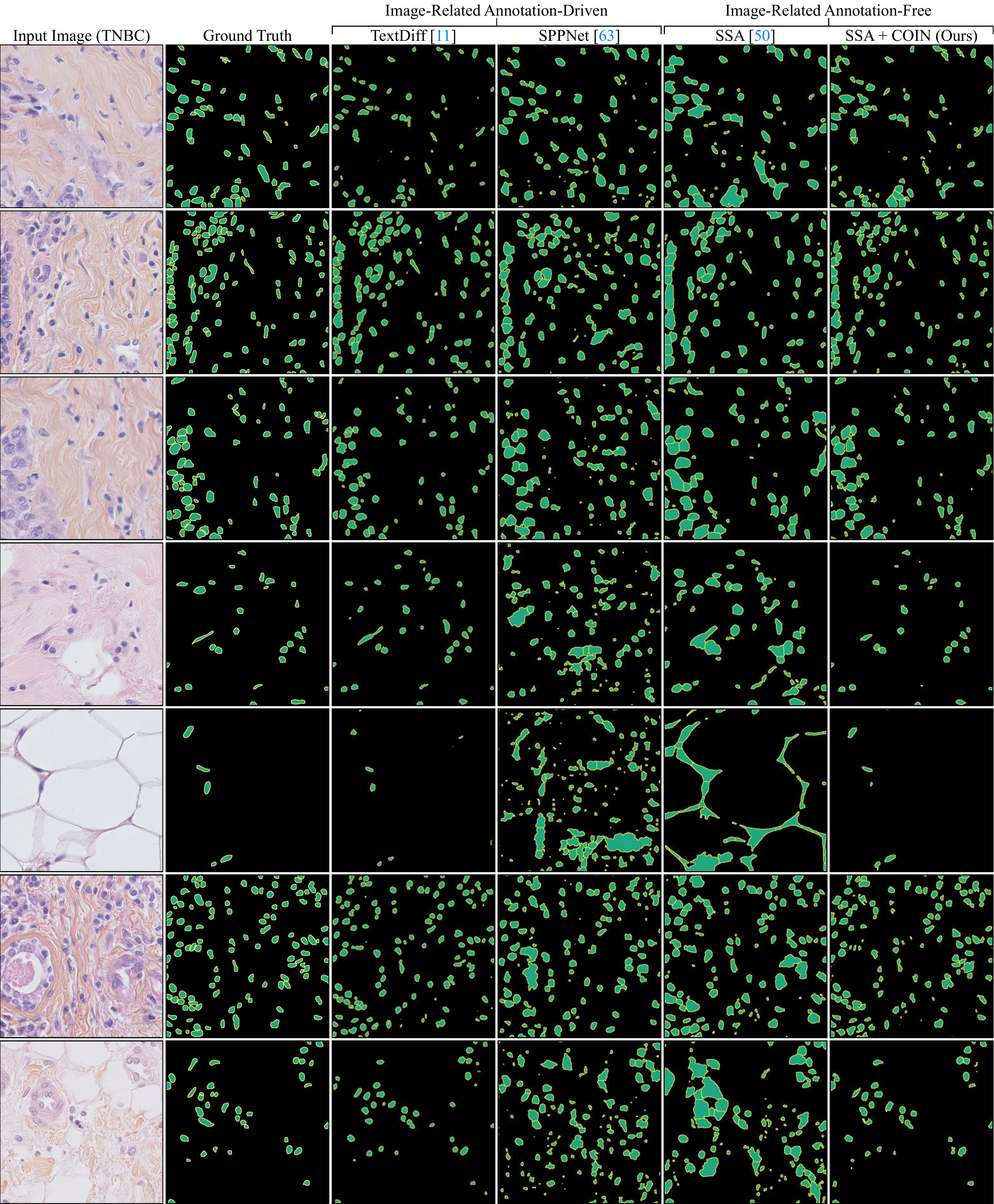}
  \caption{Qualitative comparison of annotation-driven and -free methods \cite{ssa2020miccai, feng2024textdiff, xu2023sppnet} on the TNBC \cite{naylor2019tnbc} test set.}
  \label{appendix_tnbc}
\end{figure*}

% Model-agnostic qualitative results (SSA, PSM)
\begin{figure*}[p]
  \centering
  \includegraphics[width=0.95\textwidth]{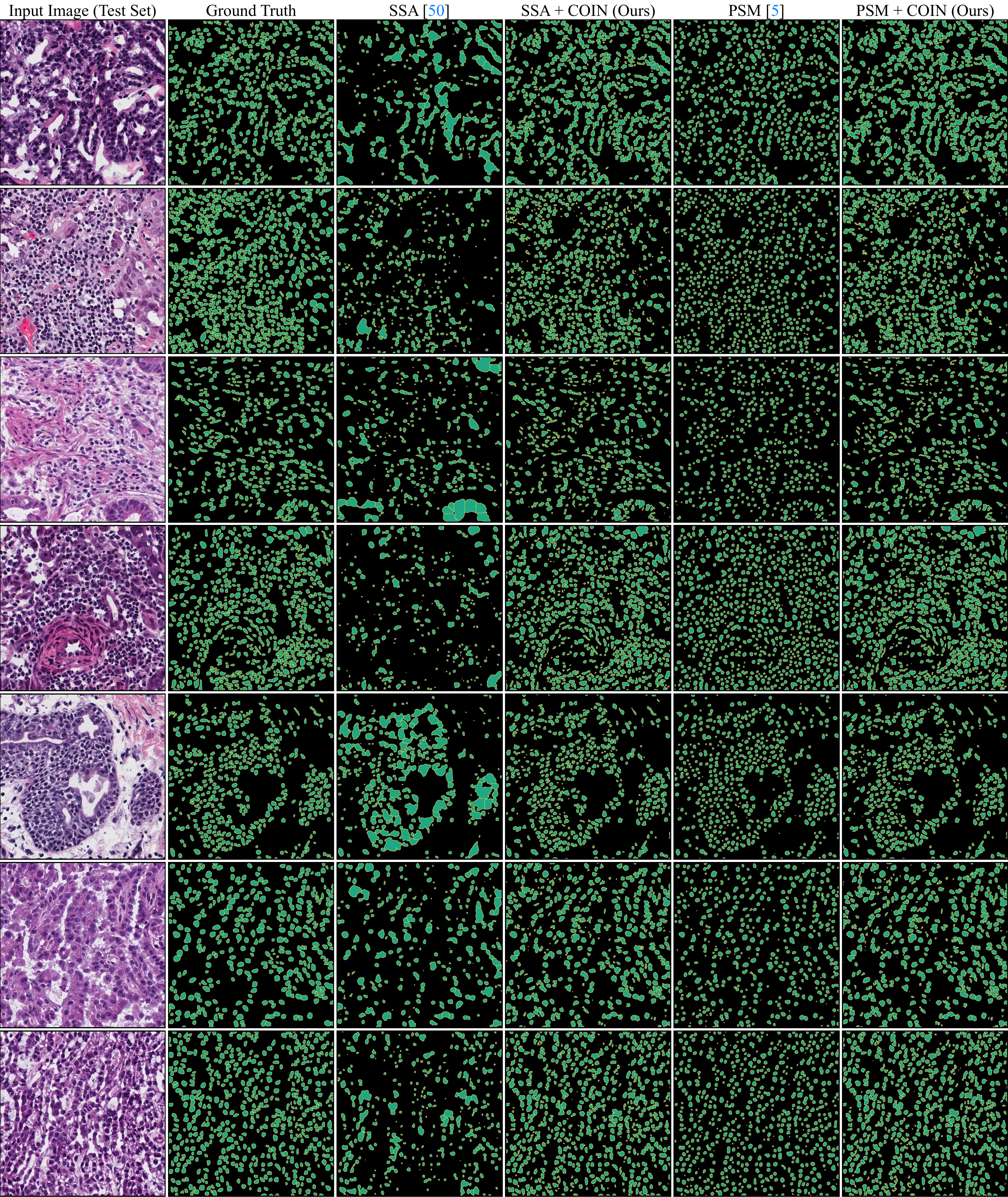}
  \caption{Model-agnostic qualitative comparison of two UCIS models \cite{ssa2020miccai, psm2023miccai} on the MoNuSeg \cite{kumar2017monuseg, kumar2020monuseg} test set.}
  \label{appendix_modelagnostic_figure}
\end{figure*}

\begin{figure*}[p]
  \centering
  \includegraphics[width=0.8\textwidth]{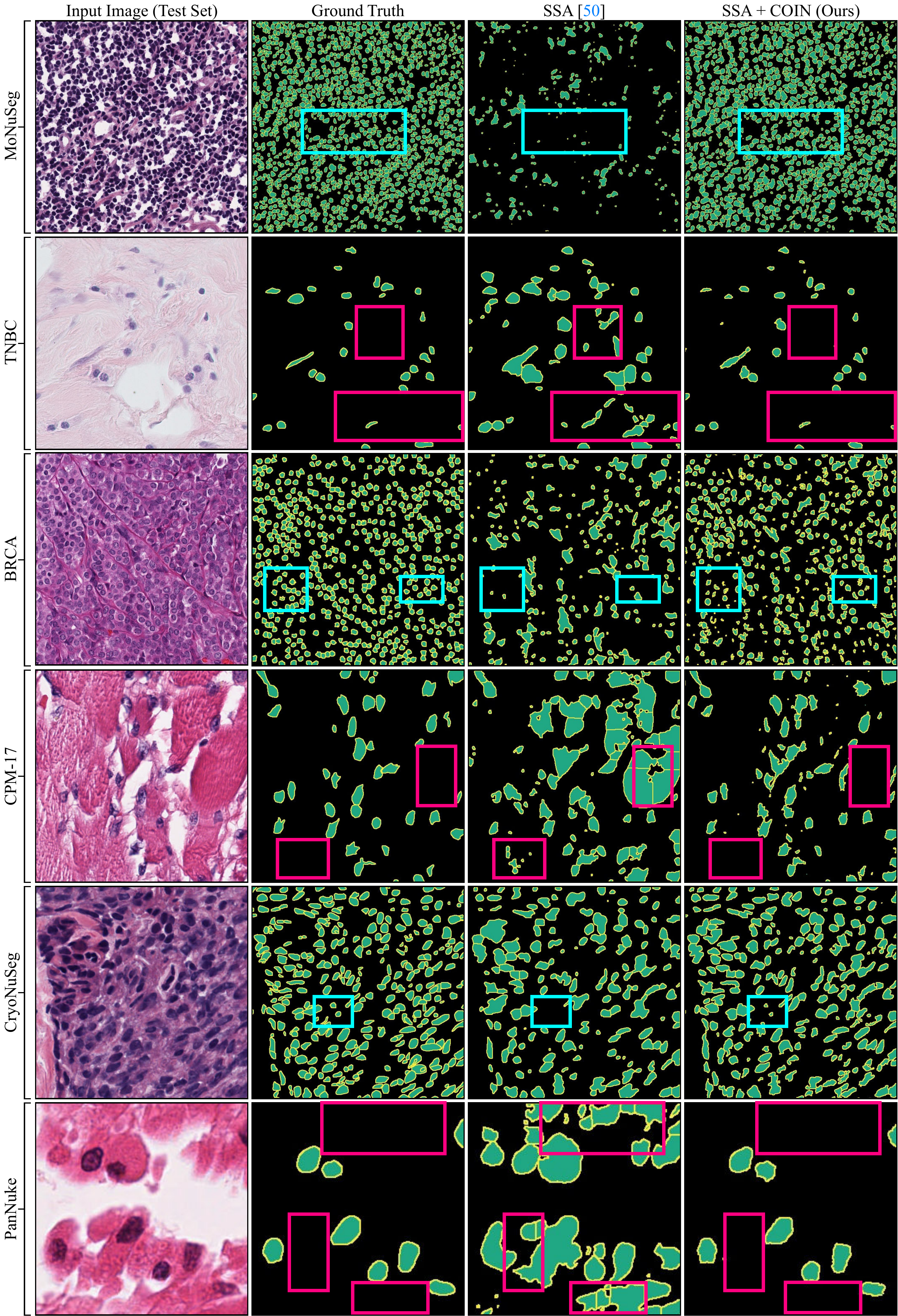}
  \caption{\textbf{Visualization of qualitative improvements on five benchmarks \cite{naylor2019tnbc, dataset2021brca, vu2018cpm, dataset2021cryonuseg, gamper2020pannuke}}. Regions marked with pink boxes represent false positives identified by the baseline \cite{ssa2020miccai}, while cyan boxes indicate false negatives.}
  \label{appendix_ModelAgnostic}
  %\vspace{-0.3cm}
\end{figure*}

% Multiple datasets qualitative results
%BRCA
\begin{figure*}[p]
  \centering
  \includegraphics[width=0.95\textwidth]{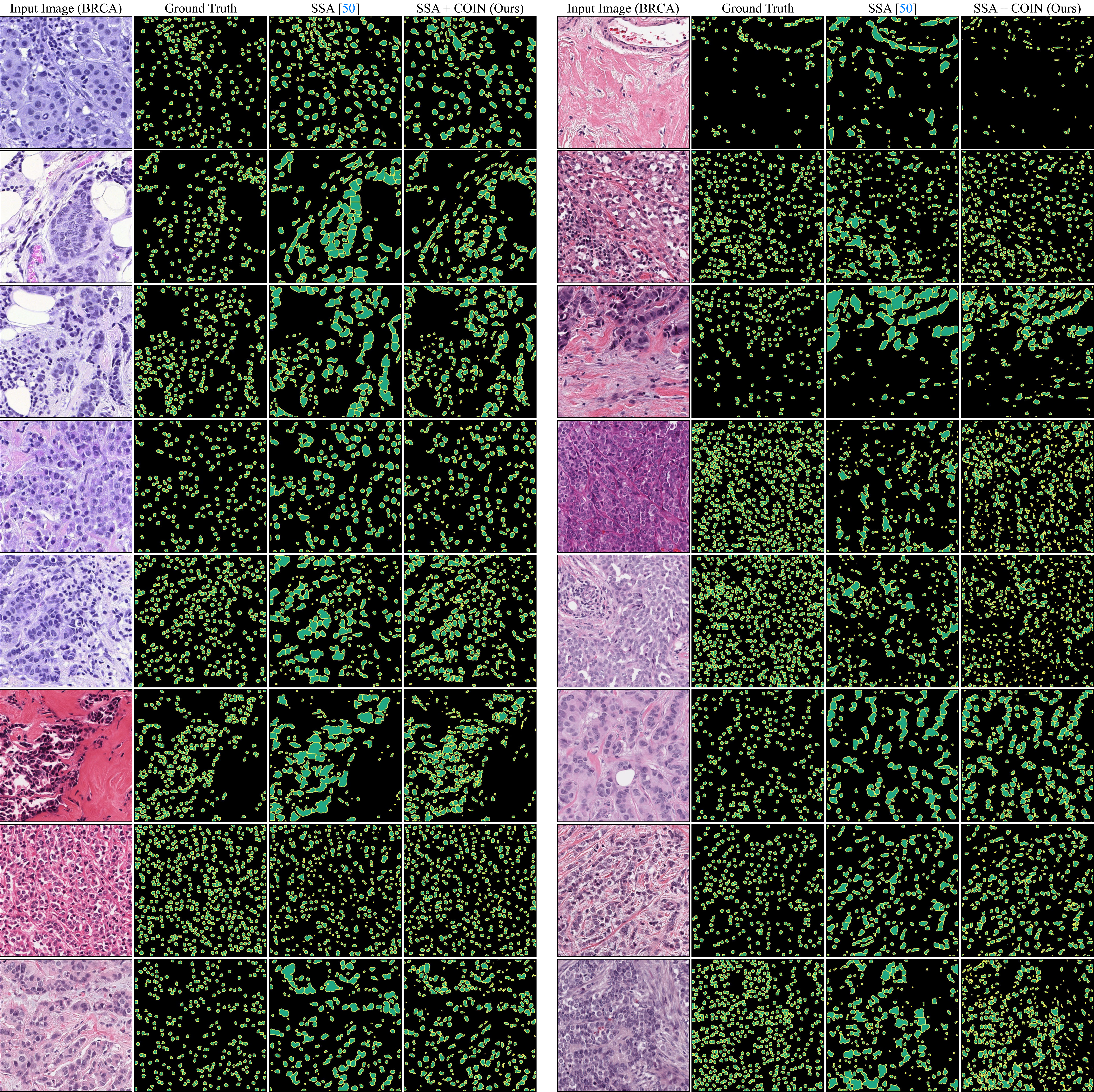}
  \caption{Qualitative examples on the BRCA \cite{dataset2021brca} test set.} 
  \label{appendix_brca}
\end{figure*}

%CPM-17
\begin{figure*}[p]
  \centering
  \includegraphics[width=0.95\textwidth]{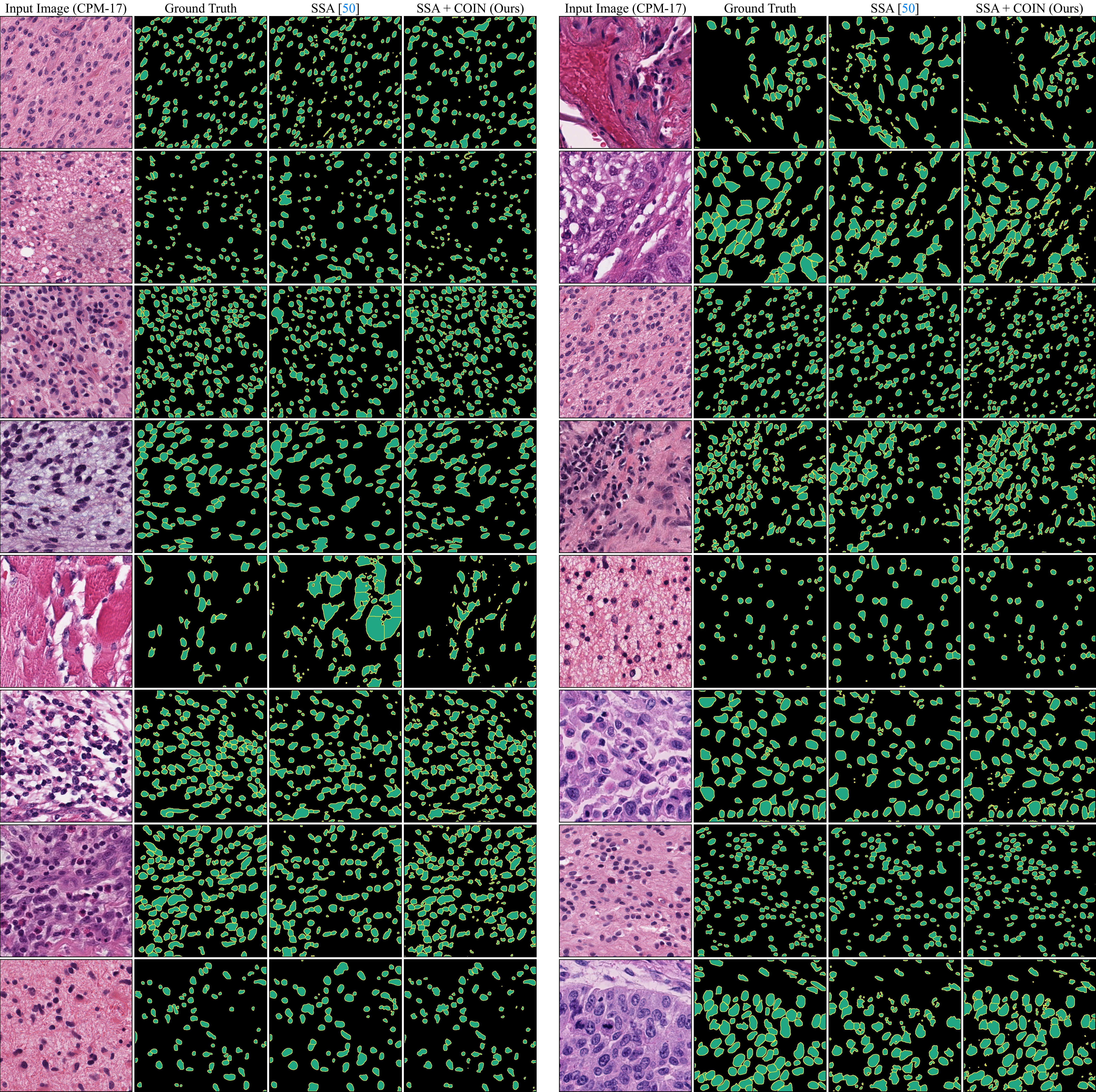}
  \caption{Qualitative examples on the CPM-17 \cite{vu2018cpm} test set.} 
  \label{appendix_cpm}
\end{figure*}

%CryoNuSeg
\begin{figure*}[p]
  \centering
  \includegraphics[width=0.7\textwidth]{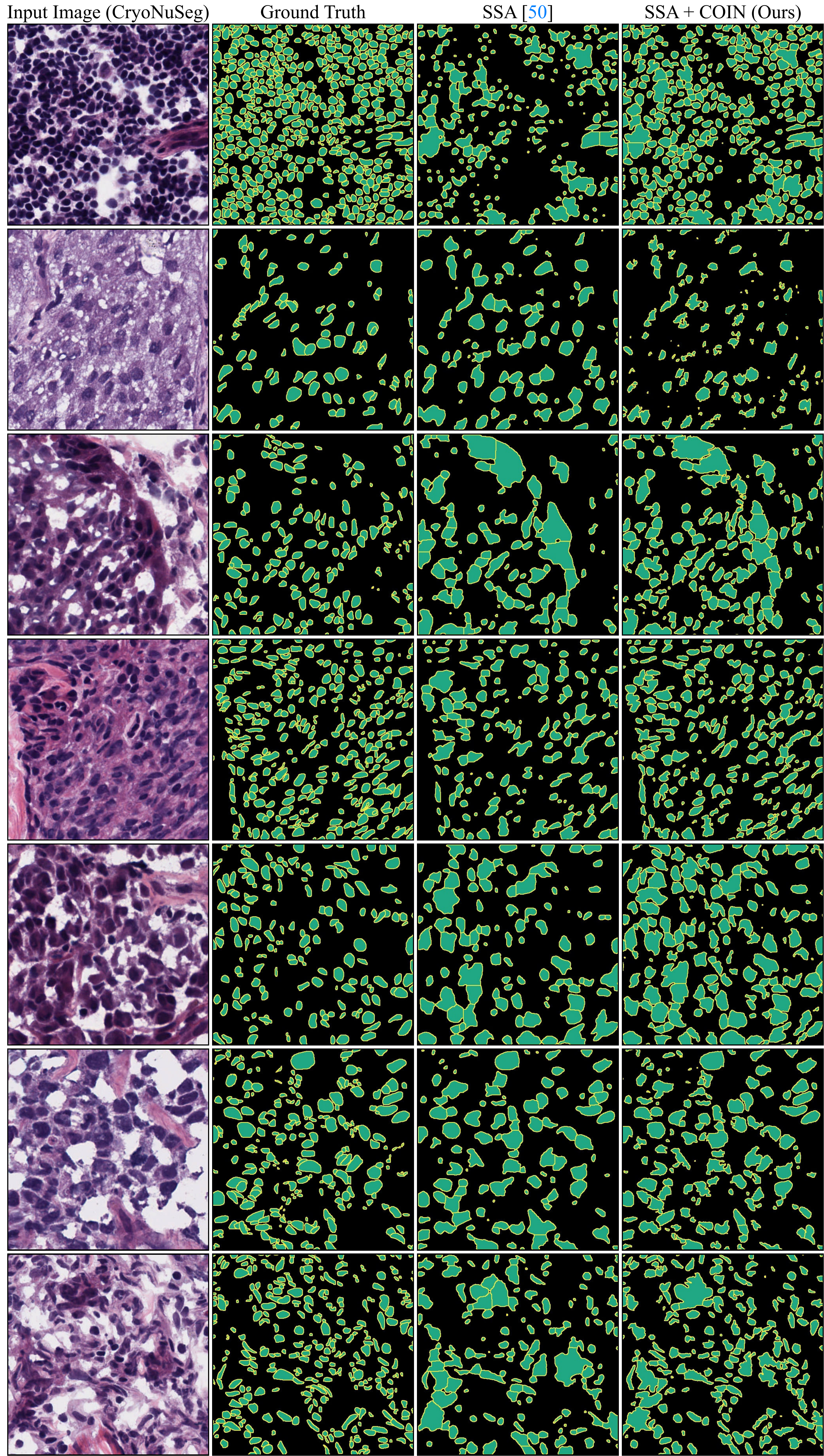}
  \caption{Qualitative examples on the CryoNuSeg \cite{dataset2021cryonuseg} test set.} 
  \label{appendix_cryonuseg}
\end{figure*}

%PanNuke
\begin{figure*}[p]
  \centering
  \includegraphics[width=0.95\textwidth]{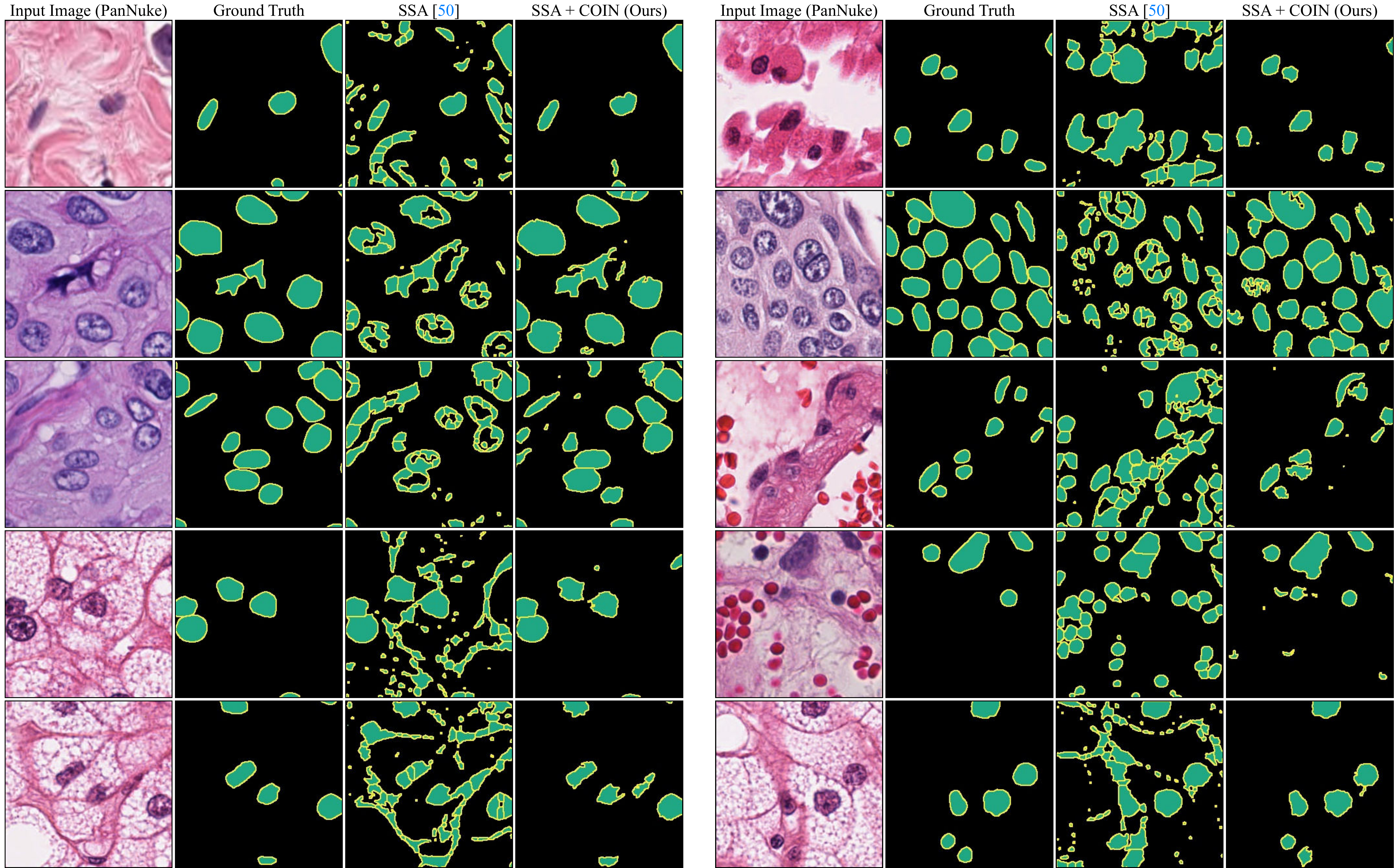}
  \caption{Qualitative examples on the PanNuke \cite{gamper2020pannuke} test set.} 
  \label{appendix_pannuke}
\end{figure*}

\end{document}